\documentclass[conference]{IEEEtran}
\IEEEoverridecommandlockouts
\usepackage{cite}
\usepackage{amsmath,amssymb,amsfonts}
\usepackage{algorithmic}
\usepackage{graphicx}
\usepackage{textcomp}
\usepackage{microtype}
\usepackage{siunitx}
\usepackage[table,xcdraw]{xcolor}
\usepackage{multirow}
\usepackage{fancyhdr}
\usepackage{array}
\usepackage{floatrow}
\usepackage{fixltx2e}
\usepackage{stfloats}
\usepackage{url}
\usepackage{times}
\usepackage{color}
\usepackage{verbatim}
\usepackage{subfiles}
\usepackage{subcaption}
\usepackage{lipsum}
\usepackage{xr}
\usepackage{nameref}
\usepackage{zref-xr}
\usepackage{adjustbox}
\usepackage{amsmath, amssymb}
\usepackage[ansinew]{inputenc}
\usepackage{mathtools}
\usepackage{esvect}
\usepackage{xspace}
\usepackage{standalone}
\usepackage{fixltx2e}
\usepackage{authblk}
\usepackage{soul}
\usepackage{booktabs}
\usepackage{multirow}
\usepackage{tabularx}
\usepackage{gensymb}
\usepackage{capt-of}
\usepackage{tcolorbox}
\usepackage{threeparttable}
\usepackage{authblk}
\usepackage{circledsteps}
\usepackage{amsmath}
\usepackage{enumitem}
\usepackage{array, caption, floatrow, tabularx, makecell, booktabs}%
\setcellgapes{3pt}
\usepackage{hyperref}
\floatstyle{plaintop}
\restylefloat{table}
\hypersetup{
    colorlinks=true,
    linkcolor=blue,
    filecolor=magenta,      
    urlcolor=cyan,
    pdftitle={Overleaf Example},
    pdfpagemode=FullScreen,
    }
\usepackage{hyperref}
\def\BibTeX{{\rm B\kern-.05em{\sc i\kern-.025em b}\kern-.08em
    T\kern-.1667em\lower.7ex\hbox{E}\kern-.125emX}}

\newcommand{\blue}[1]{\textcolor{blue}{#1}}
\newcommand{\Fig}[1]{Fig.~\ref{#1}}
\begin{document}

\fancypagestyle{firstpage}{
  \fancyhf{}
  \renewcommand{\headrulewidth}{0pt}
  \fancyhead[C]{\vspace{5pt}\normalsize{To Appear in 2022 IEEE International Symposium on Performance Analysis of Systems and Software \\ Preprint Version. Accepted April, 2022.
      \vspace{30pt}}} 
  \fancyfoot[C]{\thepage}
}

\title{Roofline Model for UAVs: A Bottleneck Analysis Tool for Onboard Compute Characterization of Autonomous Unmanned Aerial Vehicles\\
[20pt]%
\normalsize{\it ``All models are wrong, but some are useful.'' -- George Box}
}

\renewcommand\Authsep{ }
\renewcommand\Authand{  }
\renewcommand\Authands{ }

\author[$\dagger$]{Srivatsan~Krishnan}
\author[$\dagger$]{Zishen~Wan\textsuperscript{*}\thanks{\textsuperscript{*}This work was initiated and conducted while the student was at Harvard University and he is now a Ph.D. student at the Georgia Institute of Technology.}}
\author[$\mp$]{Kshitij~Bhardwaj}
\author[$\dagger$]{Ninad Jadhav}
\author[$\S$]{Aleksandra Faust}
\author[$\dagger$]{Vijay Janapa Reddi}

\setlength{\belowcaptionskip}{-10pt}

\affil[$\dagger$]{Harvard University  $^{\mp}${Lawrence Livermore National Lab}  $^\S${Google Brain Research}}

\maketitle
\thispagestyle{firstpage}
\begin{abstract}
  We introduce an early-phase bottleneck analysis and characterization model called the F-1 for designing computing systems that target autonomous Unmanned Aerial Vehicles (UAVs). The model provides insights by exploiting the fundamental relationships between various components in the autonomous UAV, such as sensor, compute, and body dynamics. To guarantee safe operation while maximizing the performance (e.g., velocity) of the UAV, the compute, sensor, and other mechanical properties must be carefully selected or designed. The F-1 model provides visual insights that can aid a system architect in understanding the optimal compute design or selection for autonomous UAVs. The model is experimentally validated using real UAVs, and the error is between 5.1\% to 9.5\% compared to real-world flight tests. An interactive web-based tool for the F-1 model called Skyline is available for free of cost use at: ~\url{https://bit.ly/skyline-tool}
\end{abstract}

\section{Introduction}

Autonomous machines like Unmanned Aerial Vehicles (UAVs) are on the rise~\cite{Timothy2017,medical-org,search-and-rescue-org,8373043,wan2021survey,liu2021robotic}. Amongst these, quadcopters account for the vast majority of the market~\cite{quad-marketshare-1, quad-marketshare-2} due to their ability to vertically take off and land (VTOL) and navigate in confined spaces. These unique capabilities enable them to be deployed in numerous applications, such as search and rescue~\cite{search-and-rescue,search-and-rescue-org}, package delivery~\cite{package-org-1}, surveillance~\cite{surveillance-org}, and sports photography~\cite{photographyh-org}. These applications have motivated the industry to build domain-specific hardware~\cite{agx-power,TX2,movidius-drone,wan2021energy}.

However, building computing for UAVs is challenging as UAVs differ from traditional systems (embedded systems, servers, etc.). They are severely size, weight, and power (SWaP) constrained. Moreover, UAVs have many components like sensors, compute mechanical frames, and rotors, interacting to work as one coherent system. The selection of these components greatly affects the UAV's velocity, mission time, energy, etc. Components such as the sensor and autonomy algorithm processing rate determine how fast the UAV reacts in a dynamic environment. Likewise, the payload weight and thrust determine if the physics allows UAVs to accelerate and move faster.

To guide compute design for SWaP and cost constrained systems, we need tools to understand the interactions between the various components and quantify the bottlenecks. UAVs are complex autonomous systems, and the computing platform is just one among many other components. To design the balanced onboard compute, we need to consider the role of computing in the context of the whole UAV system. As we demonstrate later, traditional isolated compute performance metrics can lead to misguided conclusions.

\begin{figure}[t]
\vspace{5pt}
        \includegraphics[width=\columnwidth]{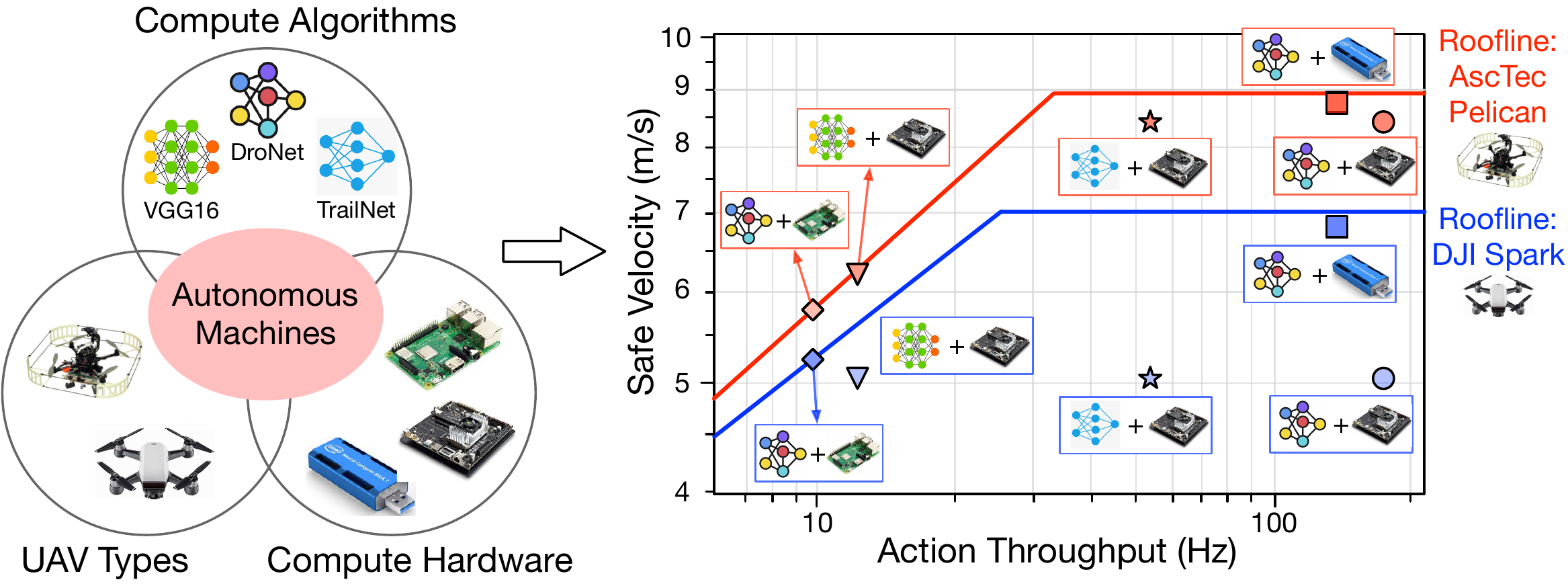}
        \vspace{-10pt}
        \caption{A bottleneck analysis tool for UAV onboard compute characterization. The tool takes a full-system view of UAVs and identifies the various bottlenecks using the F-1 model. The ``roofline'' helps us quickly determine whether the UAV is sensor bound, compute bound, or body-dynamics bound.
        \vspace{-10pt}}
        \label{fig:intro_overview}
\end{figure}

In this paper, we introduce a ``Roofline Model for UAVs,'' an early-phase bottleneck analysis tool to guide the design of balanced computing systems for UAVs. The tool determines which of the UAV components (compute, sensor, or physics) limits the safe operating velocity; \textit{safe high-speed autonomous navigation remains one of the key challenges in enabling aerial robot applications~\cite{darpa,vijaykumar-1,vijaykumar-2,droneracing}}. A high safe velocity ensures that the UAV is reactive to a dynamic environment and ensures that the UAV finishes tasks quickly, thereby lowering mission time and energy~\cite{mavbench}. \Fig{fig:intro_overview} shows that the tool takes the autonomy algorithm, onboard compute, and the UAV type to perform bottleneck analysis. The coupling between the different choices leads to different ``rooflines.'' For example, the AscTec Pelican~\cite{high-speed-drone} has a different roofline from DJI Spark~\cite{dji-spark} where the $y$-axis is the maximum safe velocity, and the $x$-axis is the decision-making rate (action throughput). The individual design points on the chart reflect the maximum safe velocity for a given UAV system configuration. The output visually resembles that of a traditional computer system roofline model~\cite{roofline}; however, unlike the roofline model, the parameters in our model quantify the UAV as a holistic system as opposed to the computing system in isolation. We can use our tool in two ways. It can be used as a visual performance model to understand various bounds and bottlenecks in the UAV system. It can also guide us toward building a balanced compute system for UAVs.

We use the tool to answer a wide array of important characterization questions for designing onboard computing systems for UAVs. First, how does the ad-hoc selection of onboard compute affect the UAV's safe velocity? Second, given a UAV and onboard compute, how do we systematically evaluate different autonomy algorithms and their impact on UAV's safe velocity? Third, given a UAV, autonomy algorithm, and sensor, how do we systematically characterize the impact of redundancy~\cite{redundancy-2} in onboard compute on UAV's safe velocity? Finally, given several onboard compute, autonomy algorithms, and sensor choices, how can we systematically select off-the-shelf components that maximize the UAV's safe velocity, and how does this selection differ as we change the UAV?  

Our deep introspection and characterization studies reveal an intertwined relationship between various components in UAVs and their mission performance. The current state-of-the-art techniques rely on selecting or designing onboard compute in an \emph{ad-hoc} fashion or based on isolated compute metrics such as peak compute throughput and low-power without considering its actual impact on the UAV's performance. For instance, our model shows that selecting onboard compute in this fashion results in 2.3$\times$ degradation in safe velocity. 

Our study also reveals that ad-hoc design choices solely based on isolated compute metrics such as throughput, power, or energy efficiency can be misleading and shows architects how much optimization effort is needed to maximize the UAVs' performance. For instance, hardware accelerators optimized solely for low-power~\cite{pulp-dronet} can degrade the safe velocity by 4.3$\times$ for a nano-UAV~\cite{nano-uav} due to their inability to make fast-enough decisions. To this end, we see a clear need for a systematic methodology to characterize onboard computers in these complex UAV systems. Our work introduces a modeling tool that gives insights into how component selections impact the UAV's safe velocity and offers optimization targets for designing balanced onboard compute for autonomous UAVs.

\begin{figure*}[ht!]
\centering
\begin{subfigure}{0.32\textwidth}
        \includegraphics[height=1.15in]{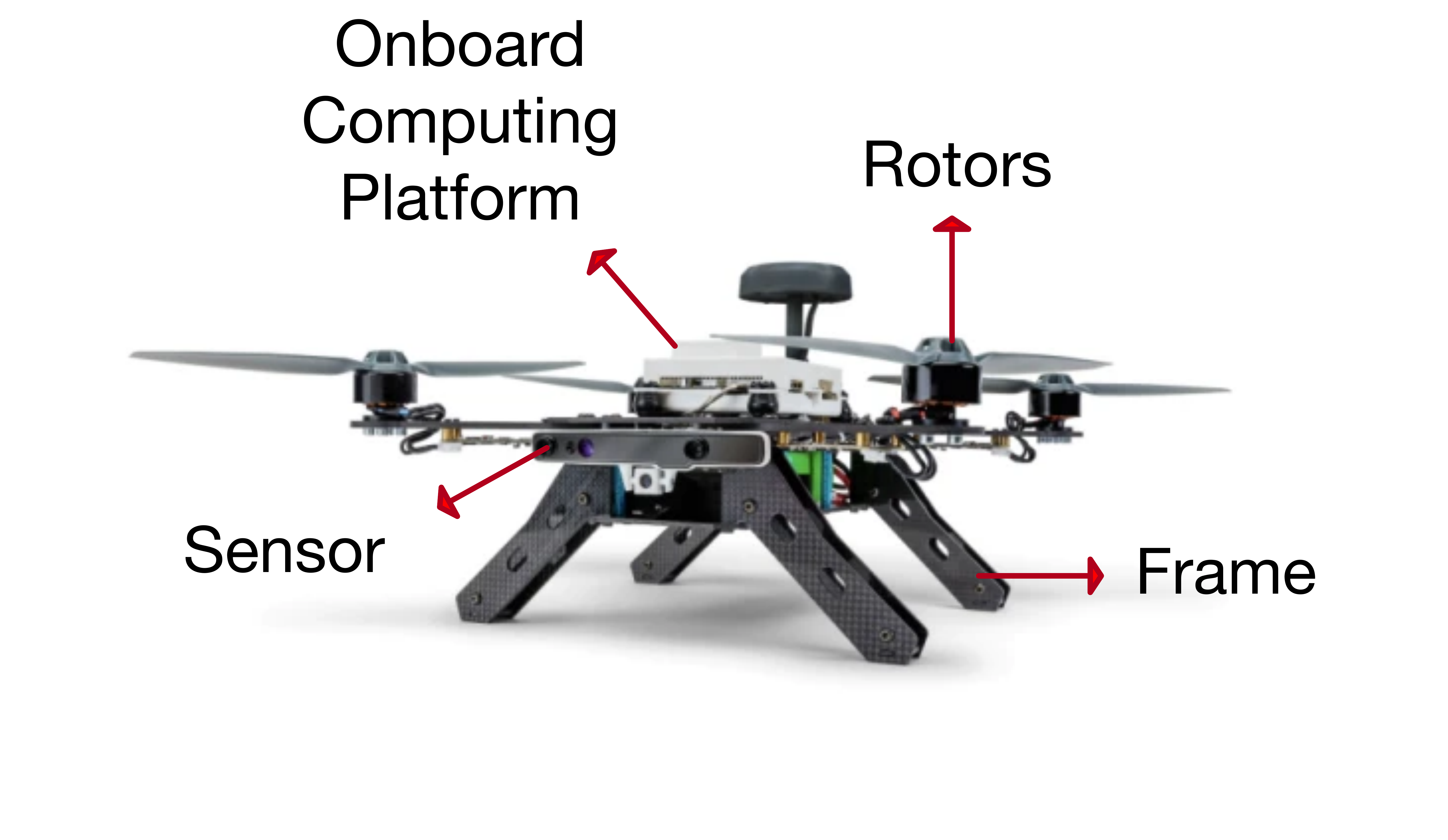}
        \caption{UAV components.}
        \label{fig:aerial-robot-comp}
        \end{subfigure}
\begin{subfigure}{0.32\textwidth}
    \includegraphics[height=1.15in]{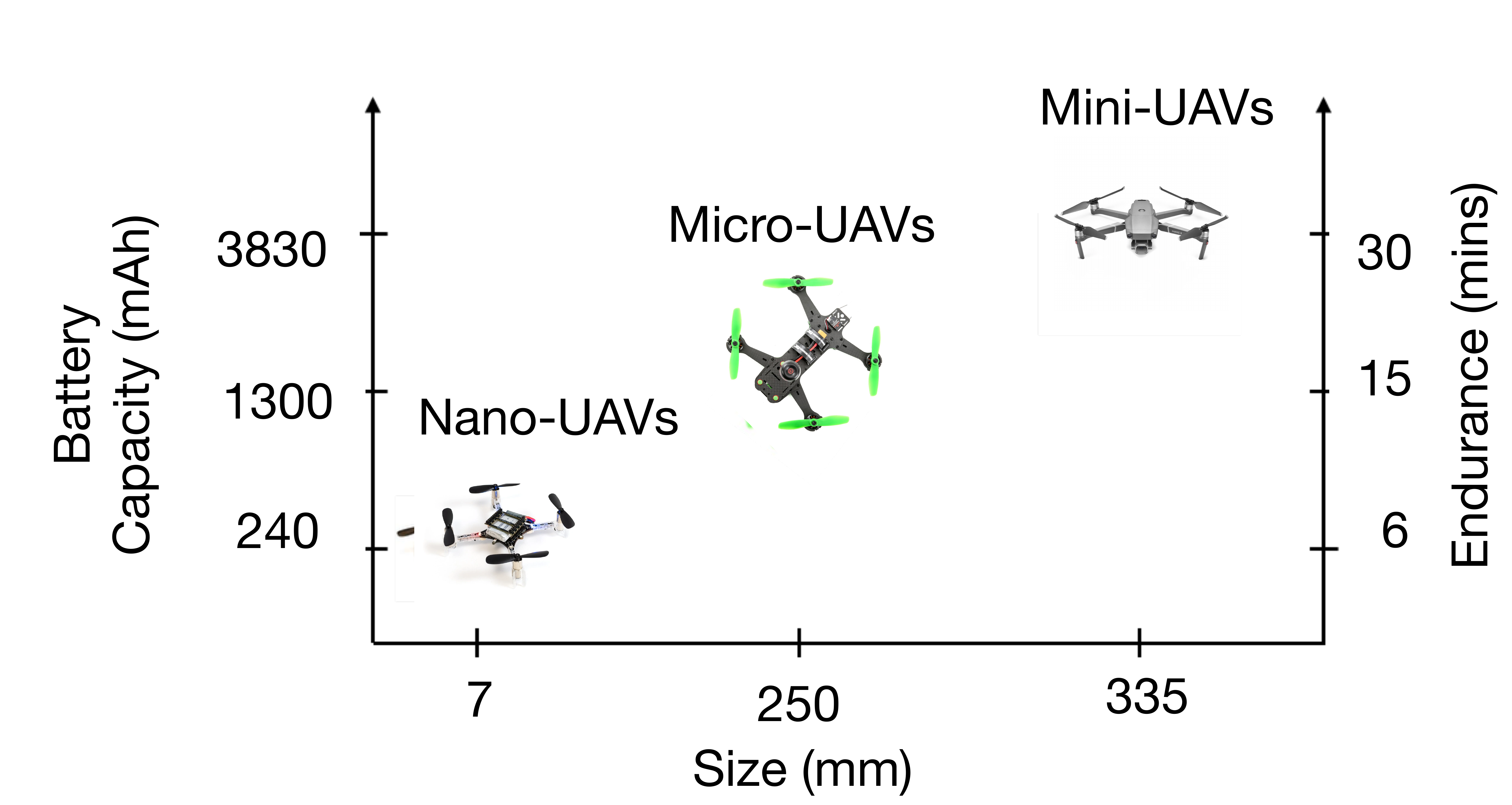}
    \caption{Size and battery capacity in UAVs.}
    \label{fig:uav-size}
    \end{subfigure}
    \hspace{10pt}
\begin{subfigure}{0.32\textwidth}
    \includegraphics[height=1.15in]{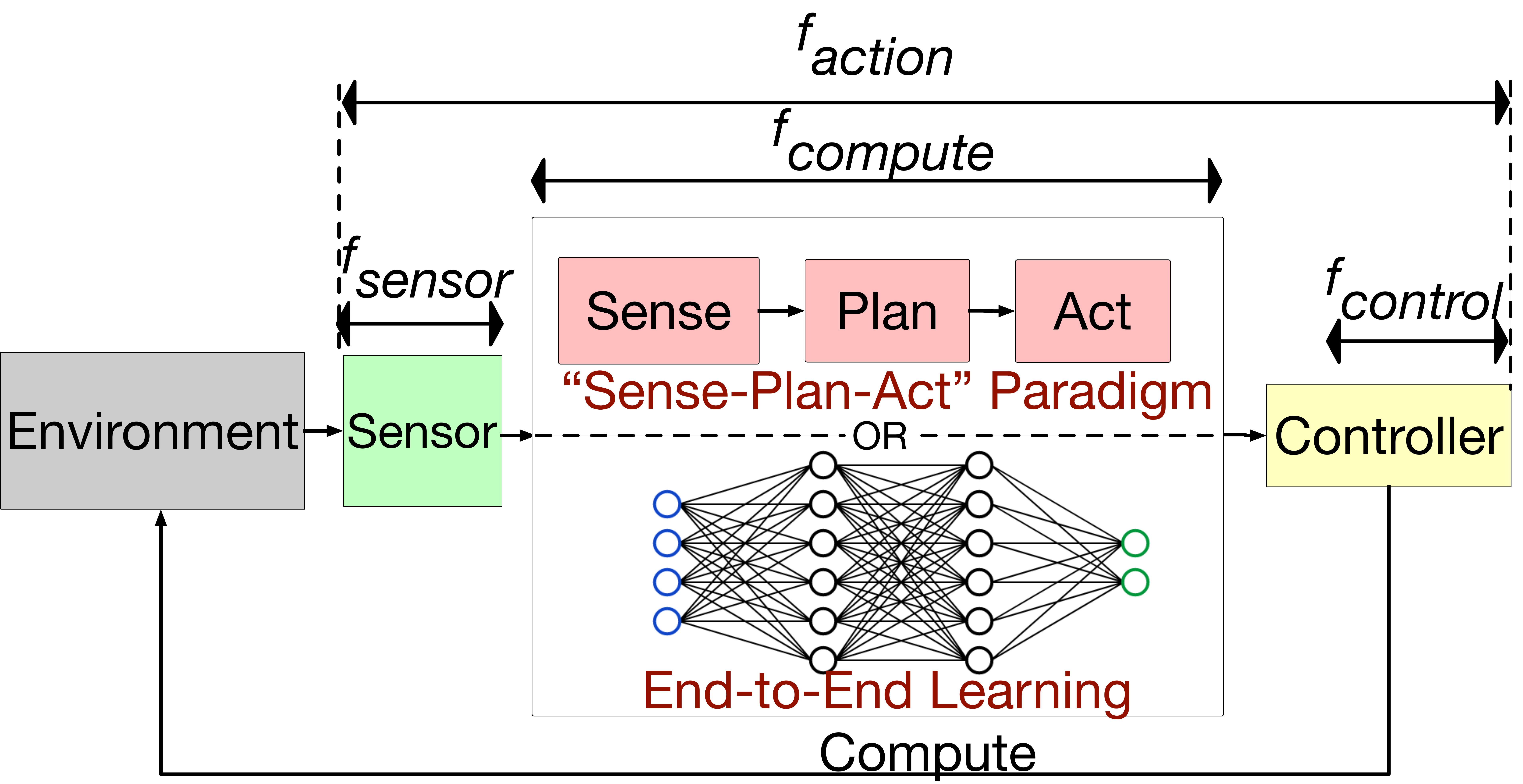}
    \caption{Autonomy algorithm paradigms.}
    \label{fig:control-algo-types}
    \end{subfigure}
    \label{fig:fund-relation}
  \caption{(A)-(B) UAV components, implications of size on battery and flight time. (C) The sensor-compute-control pipeline which determines the decision making rate, based on bottleneck analysis model~\cite{simple-models,hill2019gables}.\vspace{-15pt}}
\end{figure*}

In summary, we make the following contributions:
\begin{enumerate}
    \item A roofline-like performance model called the F-1 model for UAV system characterization. The F-1 model is experimentally validated with real-world UAV flight tests.
    \item An interactive web-based tool for the F-1 model called Skyline provides various insights into different bounds and bottlenecks as UAV components (parameters) change.
    \item A detailed characterization study using the Skyline tool by varying each UAV component and quantifying its effect on the overall, end-to-end UAV's mission performance.
    \item  A comprehensive system-level characterization of UAVs with commercial onboard compute choices and hardware accelerators explicitly built for autonomous UAVs. 
\end{enumerate}

\section{Autonomous UAVs}
\label{sec:background}

Our work primarily targets quadcopters; henceforth, we will refer to these systems as UAVs or drones interchangeably. This section provides a background on the UAV system components.


\subsection{UAV Components}

Autonomous UAVs typically have three key components (\Fig{fig:aerial-robot-comp}), namely rotors, sensors, and an onboard computing platform. Rotors determine the thrust a UAV can generate. The sensor allows the UAV to sense the environment. The computer executes the autonomy algorithm based on sensor data. The physical size of a UAV plays an important role in selecting the components. Also, the size of the UAV imposes constraints on the maximum weight of each component. A bigger UAV will have the capability to have multiple sensors and a powerful computing platform. In contrast, a smaller UAV will have limited sensor and computing capabilities. 


\subsection{Size, Weight, and Power (SWaP) Contraints}

The endurance and energy available to carry out missions vary drastically based on a UAV's size (\Fig{fig:uav-size}). The battery capacity and endurance are commensurate with the size of the UAV. For instance, a mini-UAV (e.g., Asctec Pelican) frame size is typically 350~mm and higher and has a battery with 3830 mAh. In contrast, nano-UAVs (e.g., CrazyFlie) have a frame size of around 7mm or less and have a battery of 240 mAh. As the battery capacity decreases, so does its endurance.

\subsection{Onboard Compute}

The SWaP (and cost) constraints have strong implications on the onboard compute capabilities. On one extreme, we have the nano-UAVs that, due to their size and weight, typically use microcontrollers such as ARM Cortex-M4~\cite{crazyflie} as the onboard computing platform. On the other end, mini-UAVs, which are bigger and have a higher payload carrying capacity, use a general-purpose onboard computer such as Intel NUC~\cite{high-speed-drone}.

\subsection{Flight Controller}
\label{sec:flight-controller}

While the primary onboard computer is responsible for making high-level decisions, the low-level task of stable flight and control is delegated to a dedicated flight controller, which is realized using PID controllers. The flight controller firmware stack is computationally light and is typically run on the microcontrollers~\cite{ucontroller-fc-1,ucontroller-fc-2} such as Arm Cortex M4~\cite{ucontroller-fc-1}. The flight controller uses the onboard sensors, such as the Inertial Measurement Unit (IMU)~\cite{imu} and GPS, to stabilize and control the UAV. To stabilize the UAV from unpredictable errors (sudden winds or damaged rotors), the inner-loop typically runs at closed-loop frequencies of up to 1 kHz~\cite{1khz-control,koch2019neuroflight}.

\subsection{Autonomy Algorithms}
\label{sec:algos}
Autonomy is achieved using different algorithms. 
The autonomy algorithms process the sensor data on the onboard computer to determine the high-level actions later translated to low-level controls by the flight controller. These algorithms can be classified into two broad and main categories, i.e., the ``Sense-Plan-Act'' (SPA) and ``End-to-End Learning'' (E2E). 

In  SPA,  the algorithm is broken into three or more distinct stages, namely ``sensing'', ``planning'', and ``control''. 
In the sensing stage, the sensor data is used to create a map~\cite{rusu20113d,elfes1989using,dissanayake2001solution,gao2021ielas} of the environment. The planning stage~\cite{rrt,motion-planning-survey} then uses the map to determine the best trajectory. Finally, the control stage uses the trajectory information, which actuates the rotor.

In E2E, the machine learning algorithm processes the input sensor information and they often use a neural network model to produce output actions directly. Unlike the SPA paradigm, the end-to-end learning methods do not require maps or separate planning stages. 
The model can be trained using supervised learning~\cite{e2e-nvidia,dronet,trailnet0,trail-net} or reinforcement learning~\cite{cad2rl,qt-opt,quarl,airlearning,anwar2020autonomous}.

\section{F-1: An Insightful Visual Performance Model}
\label{sec:f-1-roofline}
This section introduces a visual performance model, which we refer to as \textit{F-1}, that helps us understand whether a UAV's performance is bottlenecked by onboard compute or by other components such as sensors or UAV physics.\footnote{The F-1 name is inspired by the high-performance Formula-1 (F-1) racing cars, where components are added, removed, engineered and sometimes even fine-tuned based on the vehicle's aerodynamic regulations and specifications.} We start with the F-1 model overview and explain how it can be useful. We then describe how we construct the F-1 model.

\subsection{A Visual Performance Model}
\label{sec:f1_model}

\begin{figure}[t!]
\centering
        \begin{subfigure}{0.37\linewidth}
        \includegraphics[width=\columnwidth, keepaspectratio]{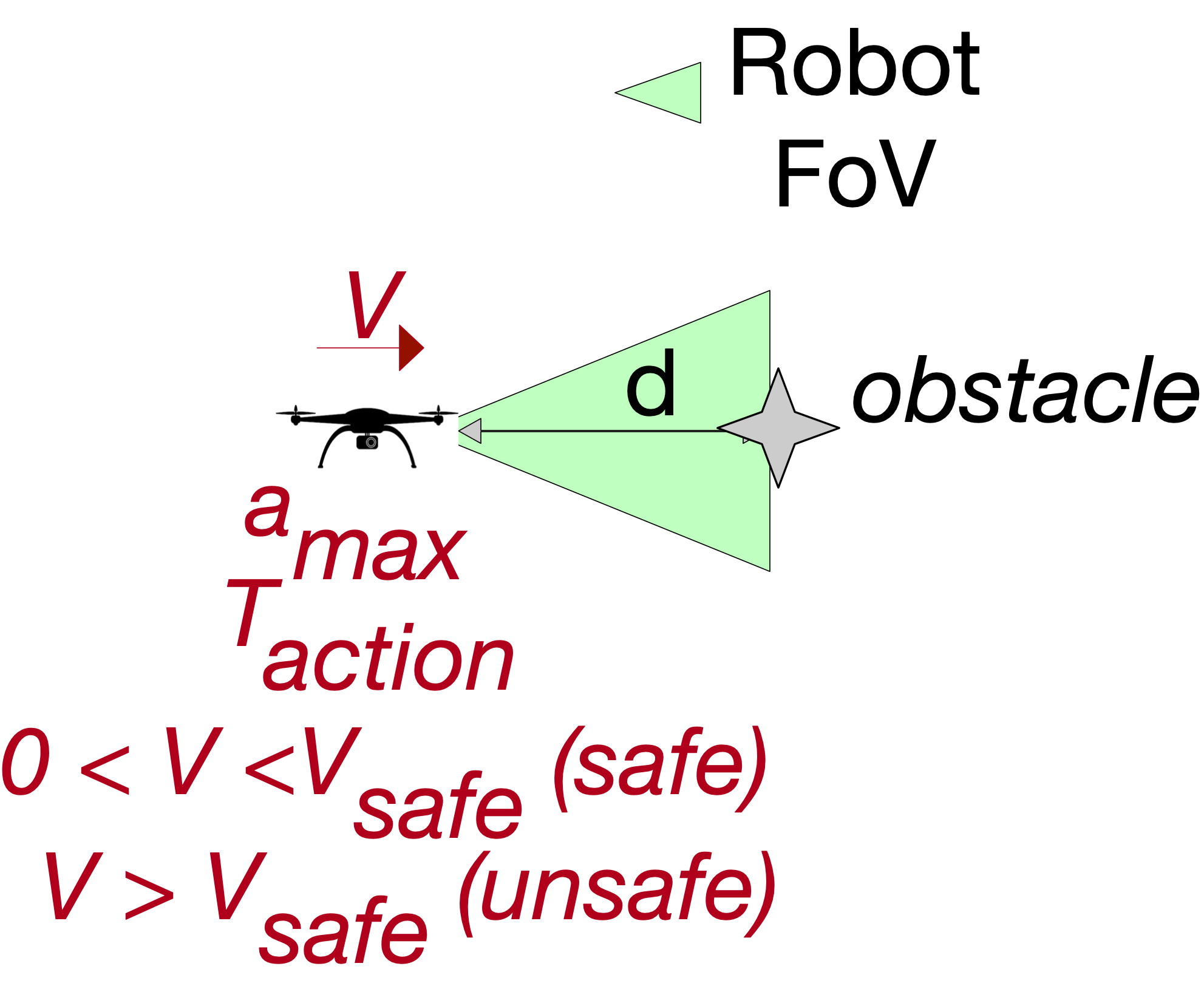}
        \caption{\footnotesize Safety Model.}
        \label{fig:safety-model}
        \end{subfigure}
        \begin{subfigure}{0.6\linewidth}
        \includegraphics[width=0.8\columnwidth, keepaspectratio]{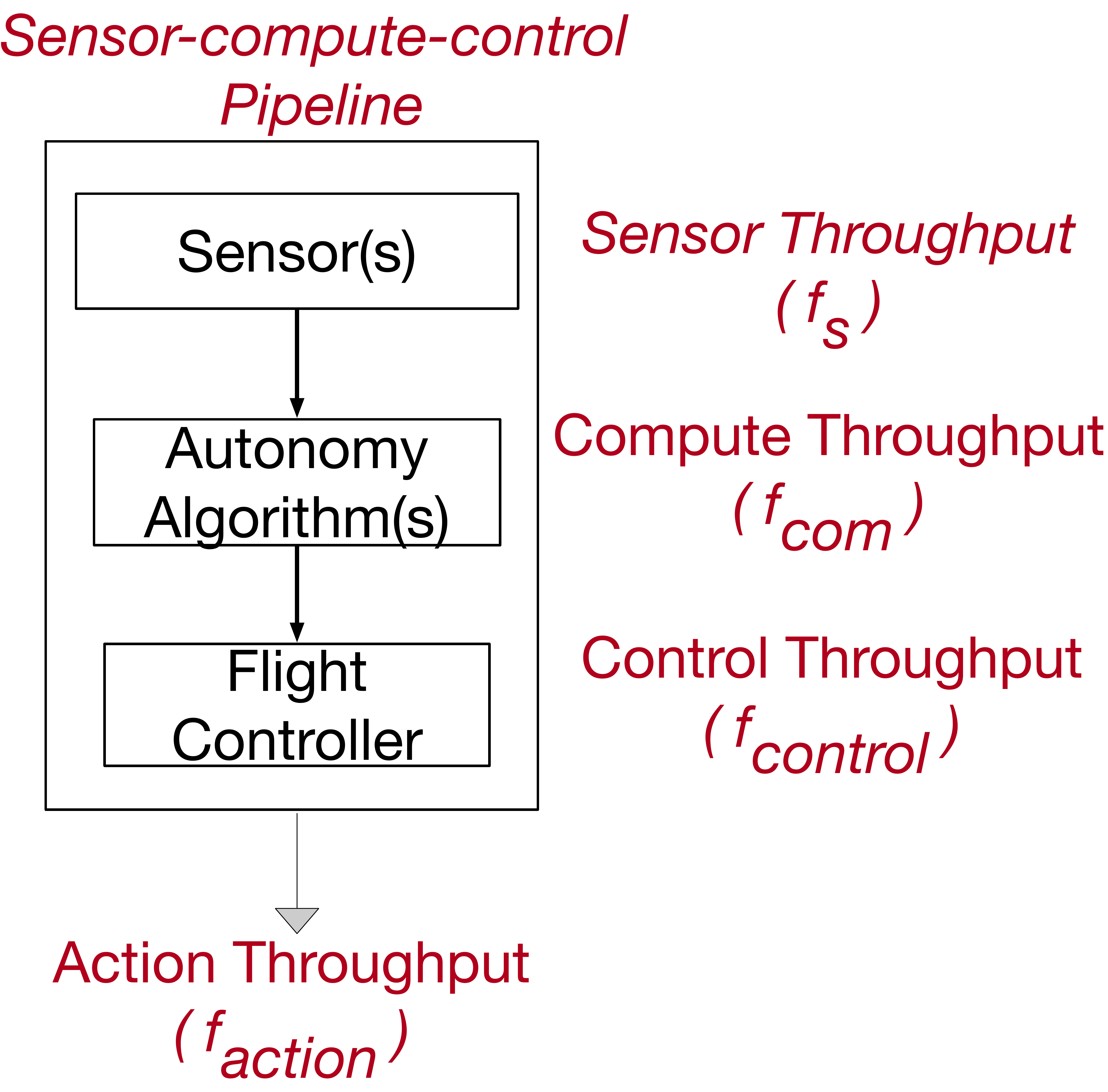}
        \caption{Sensor-compute-control pipeline.}
        \label{fig:pipeline}
        \end{subfigure}
    \label{fig:fund-relation}
  \caption{(A) The safety model for UAV. (B) Action throughput in Sensor-compute-control pipeline.}
\setlength{\belowcaptionskip}{-20pt}
\end{figure}

The F-1 model establishes a roofline-like relationship between the safe velocity of the UAV and its decision-making rate. To intuitively understand safe velocity, let us consider a simple scenario depicted in  \Fig{fig:safety-model} where a UAV has a sensor with a sensing distance of `d' meters and a field of view (FoV)~\cite{fov}. The UAV can sense any obstacle within the FoV, and the autonomy algorithm running on an onboard computer can make appropriate decisions at the rate of f$_{action}$. Considering all the payload (sensor, compute, battery, etc.), the UAV's physics can allow it to accelerate by up to a$_{max}$. In such cases, the UAV can safely fly up to a velocity of V$_{safe}$ such that even if an obstacle is within its FoV, it can safely stop without colliding. Any velocity greater than V$_{safe}$ means the UAV cannot stop safely and will collide with the obstacle. Achieving high safe velocity is crucial as it lowers the mission time and overall mission energy~\cite{mavbench}.

The decision-making rate (`Action Throughput') determines how fast the UAV can sense, process, and take appropriate actions. It can be calculated as the throughput of the sensor-compute-control pipeline shown in \Fig{fig:pipeline}. As the stages in the sensor-compute-control pipeline can be run concurrently, the minimum latency of the pipeline can never be smaller than the maximum latency of each component in the subsystem:
\begin{equation}
max (T_{sensor}, T_{compute}, T_{control}) \leq T_{action}
 \label{eq:t2act-min}
\end{equation}

If the stages of the pipeline are not fully overlapped, the total pipeline latency can never exceed:

\begin{equation}
T_{action} \leq T_{sensor} + T_{compute} + T_{control}
 \label{eq:t2act}
\end{equation}


Thus, between Eq.~\ref{eq:t2act-min} and Eq.~\ref{eq:t2act} we have the lower bound and upper bound in the pipeline latency. From these relationship we can estimate the upper-bound on the Action Throughput (f$_{action}$) using the bottleneck analysis~\cite{simple-models,hill2019gables}:

\begin{equation}
f_{action} = min(\frac{1}{T_{sensor}},\frac{1}{T_{compute}}, \frac{1}{T_{control}})
 \label{eq:f2act},
\end{equation}

\textbf{T$_{sensor}$} = 1/f$_{sensor}$ is the latency to sample data from the sensor. If the UAV has 60 FPS camera, the sensor data can be sampled at 16.67 ms interval, i.e., the sensor latency. 

\textbf{T$_{compute}$} = 1/f$_{compute}$ is the latency of the algorithm to estimate the high-level action commands. We can estimate the compute throughput (f$_{compute}$) by characterizing the autonomy algorithm on a given onboard compute. 



\textbf{T$_{control}$} = 1/f$_{control}$ is the latency to generate the low-level actuation commands. Typical f$_{control}$ values are $\sim$1 kHz~\cite{koch2019neuroflight}.

\begin{figure}[t!]
\centering
\hspace{-9pt}
        \begin{subfigure}{0.33\linewidth}
        \includegraphics[width=\columnwidth, keepaspectratio]{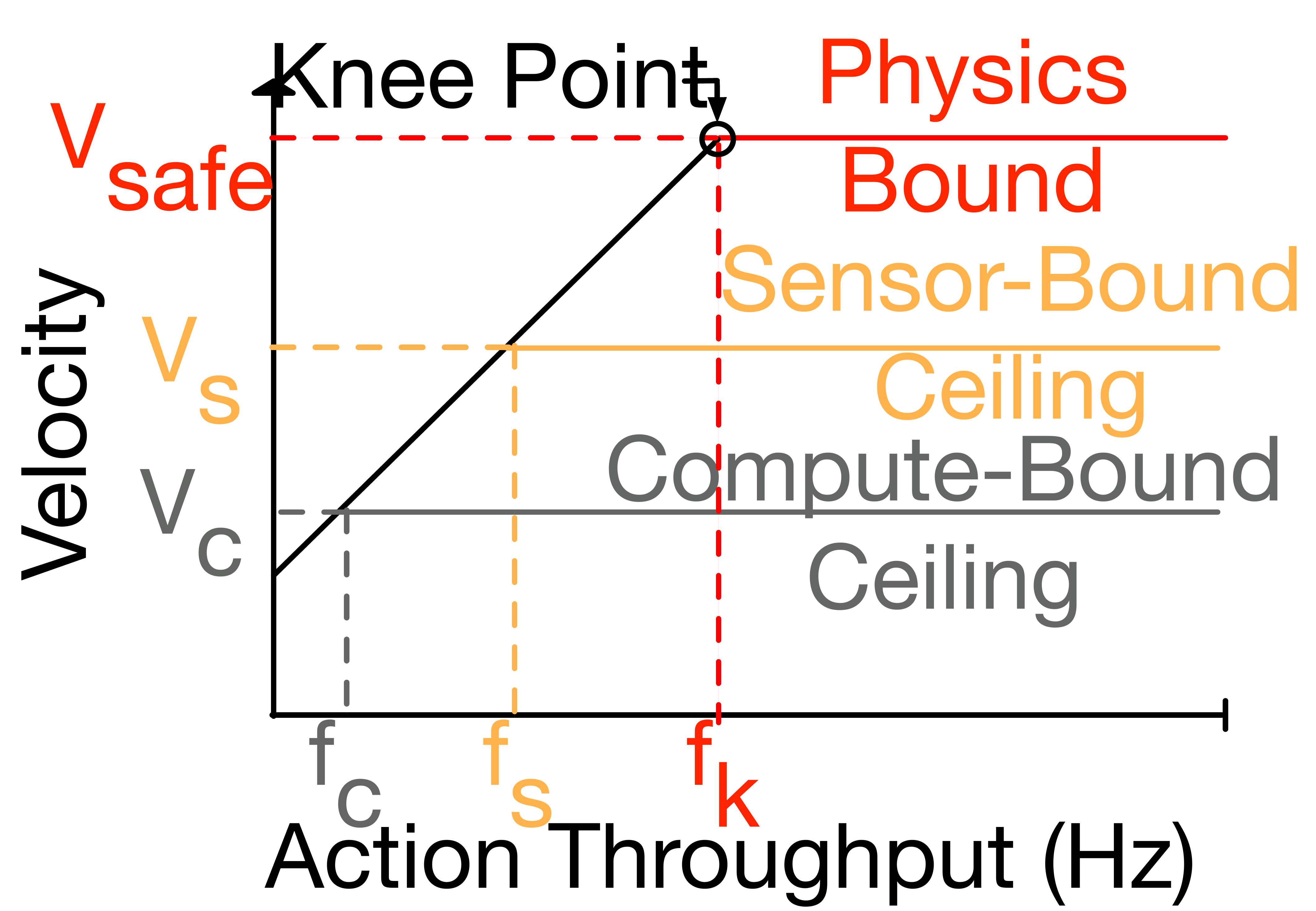}
        \caption{\footnotesize Different Bounds.}
        \label{fig:sen-bound}
        \end{subfigure}
\hspace{-7pt}
        \begin{subfigure}{0.33\linewidth}
        \includegraphics[width=\columnwidth, keepaspectratio]{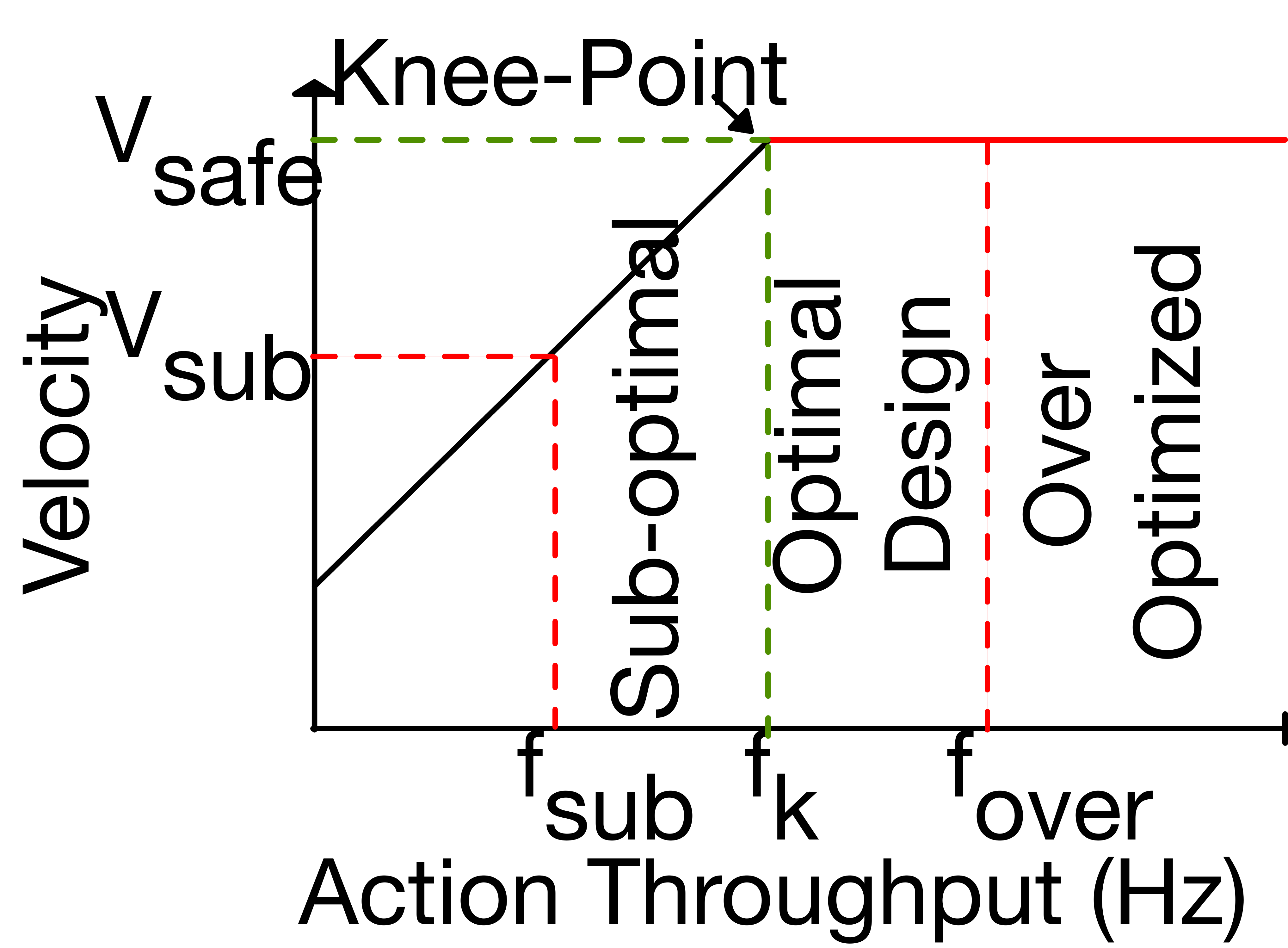}
        \caption{\footnotesize \footnotesize Optimal design.}
        \label{fig:optimal-design}
        \end{subfigure}
\hspace{-7pt}
        \begin{subfigure}{0.367\linewidth}
        \includegraphics[width=\columnwidth, keepaspectratio]{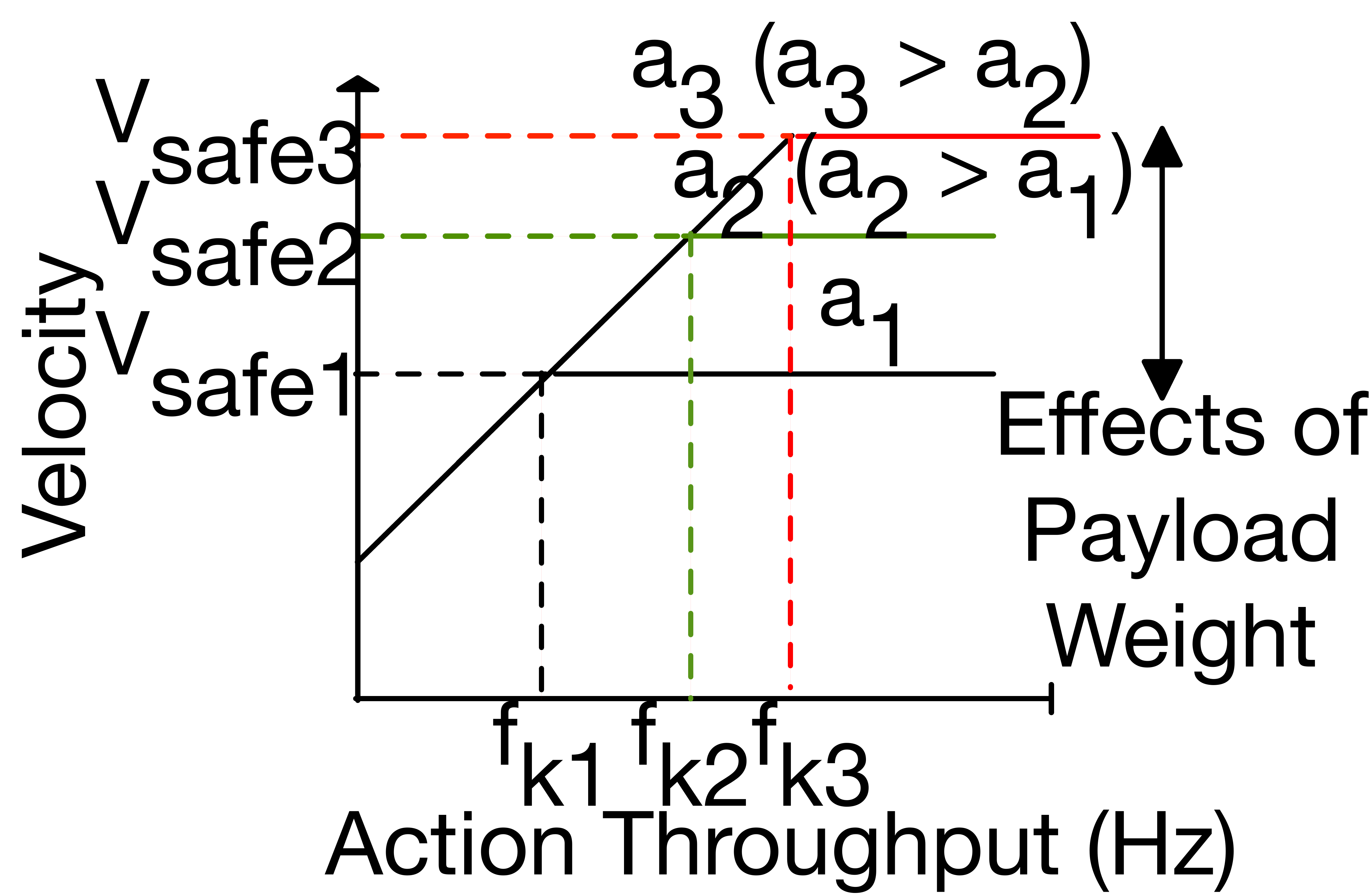}
        \caption{\footnotesize Effect of a$_{max}$.}
            \label{fig:eff-a}
        \end{subfigure}

\caption{	(A) Using the F-1 model to understand different bounds in compute, sensor, and physics, respectively. (B) Determining optimal design using the F-1 model. (C) Weight affects a$_{max}$.}
\label{fig:f1-model}
\end{figure}

The relationship between the safe velocity (V$_{safe}$) and {\em Action Throughput (f$_{action}$),} results in a roofline-like model as shown in \Fig{fig:sen-bound}. The model can be expanded to provide meaningful abstractions such as bounds and bottleneck analysis for computer architects designing onboard computers.

\subsection{Sensor, Compute and Physics Bounds}
The F-1 model can be used to perform a bound-and-bottleneck analysis to determine if the safe velocity is affected by the sensor/compute or the UAV's physics. Any point to the left of the knee-point in F-1 (\Fig{fig:sen-bound}) denotes that the safe velocity is bounded by the compute or sensor, and any point to the right of the knee-point denotes the velocity is bounded by the physics of the UAV. To achieve a balanced pipeline design, its action throughput should be equal to that of the knee-point.





\textbf{Physics Bound.} A UAV's physical properties, such as weight and thrust produced by its rotors, determine how fast it can move. Hence, the ultimate bound on the safe velocity (V$_{safe}$) will be determined by its physics (i.e., body dynamics). We call the region to the right of the knee-point (i.e., when sense-to-act throughput is greater than or equal to $f_k$) as {\em Physics bound.} Unless the physical components are improved, the velocity cannot exceed the current peak safe velocity no matter how fast a decision is made (i.e., faster compute/sensor).


\textbf{Sensor Bound.} The choice of onboard sensors may also limit the decision-making rate (f$_{action}$), which in turn can limit the safe velocity (V$_{safe}$). As shown in~\Fig{fig:sen-bound}, a robot's velocity is sensor-bound if its action throughput is equal to the sensor's frame rate ($f_{sensor}$) but less than the knee-point throughput ($f_k$). The sensor-bound case occurs when the  compute throughput (f$_{compute}$) is greater than the sensor throughput (f$_{sensor}$), and $f_{sensor}$ $<$ $f_k$. In this scenario, the sensor adds a new ceiling, thus, bounding the velocity under $V_s$. In this region, unless the sensor throughput is improved, velocity cannot exceed the sensor-bound ceiling ($V_s$) no matter how fast the onboard computer can process the sensor input.

\textbf{Compute Bound.}
The choice of onboard compute (or autonomy algorithm) also affects the decision-making rate (f$_{action}$). \Fig{fig:sen-bound} shows that a UAV's velocity is compute-bound if its compute throughput ($f_{compute}$) is less than the sensor's frame rate ($f_{s}$) and the knee-point throughput ($f_k$). The computing platform adds a new ceiling to the model, bounding the velocity under this limit ($V_c$). Unless the compute throughput is improved, the velocity cannot exceed $V_c$.

\subsection{F-1 Model for Quantifying Optimal Compute Designs}

Every UAV configuration has a unique F-1 model, thus resulting in a unique knee-point. Recall that the knee-point is the minimum action throughput required to maximize the safe velocity. Therefore, we can use this information to determine an ideal or balanced onboard compute (or autonomy algorithm) for a given UAV. Furthermore, if the compute system design is sub-optimal, it helps us understand the performance gap between the current compute design and the optimal design.

\textbf{Optimal Design.} 
For a given UAV with fixed mechanical properties, changing the sensor, onboard compute, or autonomy algorithm affects the f$_{action}$. Consequently, the optimal design point is when the action throughput is equal to the knee-point throughput (\emph{$f_k$}) as shown in \Fig{fig:optimal-design}.

\textbf{Over-Optimal Design.} If the action throughput \emph{f$_{over}$}~\emph{$>~$f$_{k}$}, then either the sensor/computer is over-optimized since any value greater than \emph{f$_{k}$} yields no improvement in the velocity of the UAV. Such an over-designed computing/sensor involves extra optimization effort. 

\textbf{Sub-Optimal Design.} If the action throughput is \emph{f$_{sub}$}, such that \emph{f$_{sub}$}~$<$~\emph{f$_{k}$}, then the sensor/computer is under-optimized, which signifies that the system if off by (\emph{f$_{sub}$}~$-$~\emph{f$_{k}$}) and there is scope for improvement through a better algorithm or selection (or design) of the computing system as shown in \Fig{fig:optimal-design}. 

\begin{figure}[t!]
\centering
        \begin{subfigure}{0.47\linewidth}
        \includegraphics[height=1in, keepaspectratio]{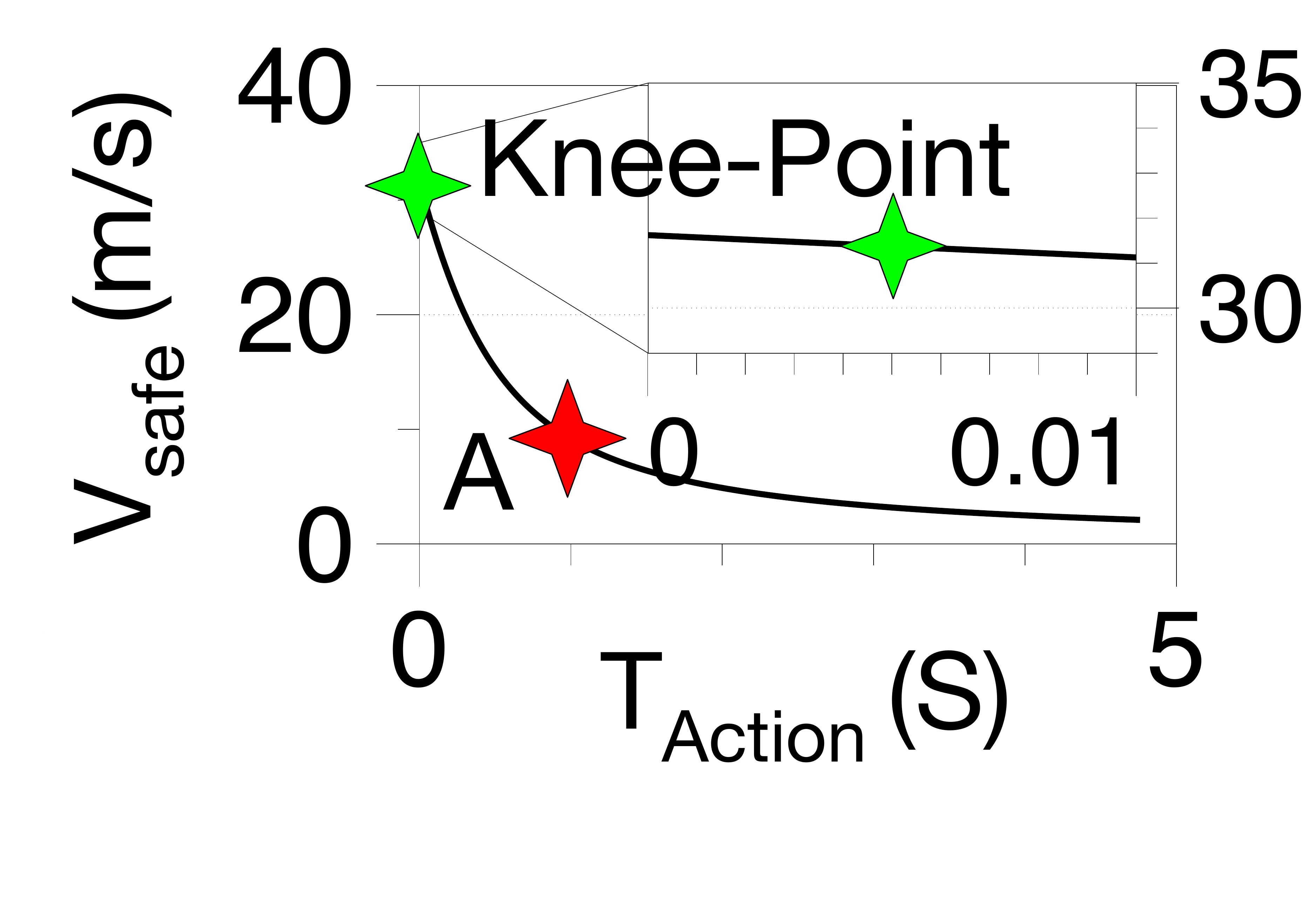}
        \caption{\footnotesize Safety Model.}
        \label{fig:cps-plot}
        \end{subfigure}
        \hfill
        \begin{subfigure}{0.47\linewidth}
        \includegraphics[height=1in, keepaspectratio]{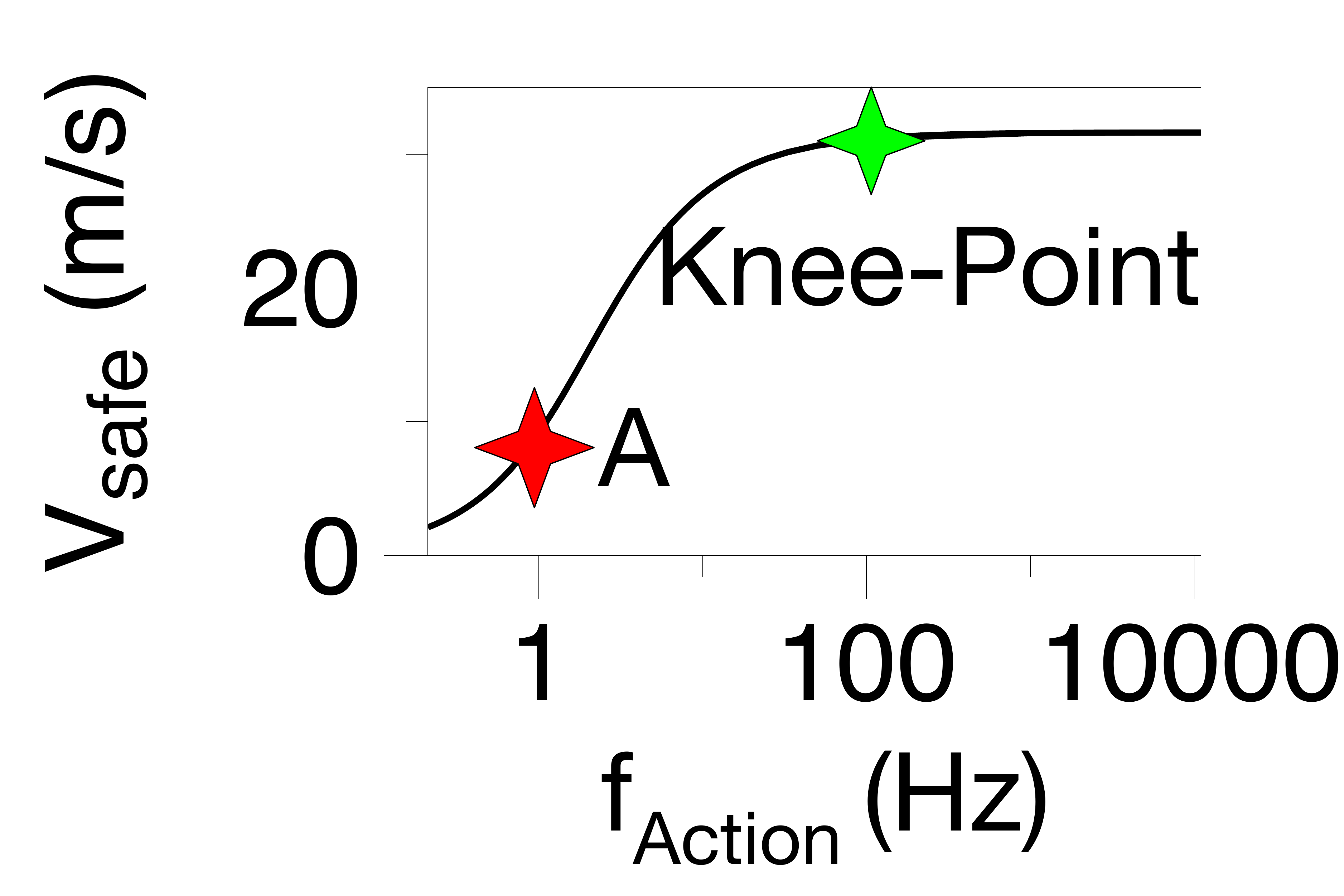}
        \caption{\footnotesize F-1 Plot.}
        \label{fig:roofline-drone-logx}
        \end{subfigure}
    \label{fig:fund-relation}
  \caption{The safety model and the F-1 roofline model.}
\end{figure}

\subsection{Constructing the F-1 Visual Performance Model}
\label{sec:f-1_construct}


In this section, we describe how we construct the F-1 model starting from prior work~\cite{high-speed-drone} that has established and validated the relationship between the UAV component parameters and the safe velocity of a UAV as described by the equation below:

\begin{equation}
    v_{safe} = \text{a}_{max} (\sqrt{T_{action}^2 + 2 \frac{d}{\text{a}_{max}}}-T_{action})
     \label{eq:vmax}
\end{equation}

Eq.~\ref{eq:vmax} states that if the UAVs physics permits it to accelerate by a$_{max}$, its compute and sensors permit it to sense and act at T$_{action}$ (1/f$_{action}$), and its sensor(s) can sense the environment as far as `d' meters, then robot can travel as fast as V$_{safe}$. 

To construct the model, we sweep the T$_{action}$ from 0 $\rightarrow$ 5~\textit{s} along with typical accelerations values (a$_{max}$ = 50 $m/s^2$) and the sensor range (\textit{d} = 10 m), as shown in~\Fig{fig:cps-plot}. We observe an asymptotic relation between velocity and T$_{action}$ such that as T$_{action}$ $\rightarrow$ 0, the velocity $\rightarrow$ 32 (\Fig{fig:roofline-drone-logx}). Likewise, as the T$_{action}$ $\rightarrow$ $\infty$, the velocity $\rightarrow$ 0. To derive F-1 from Figure~\ref{fig:cps-plot}, we plot the f$_{action }$ (inverse of T$_{action}$) and velocity in \Fig{fig:roofline-drone-logx}. 
There is a point beyond which increasing f$_{action}$ does not increase the velocity, showing a {\em roofline.} 
We annotate the plots with two sample points denoted as point `A' and `knee-point.' The point A has an f$_{action}$ of 1~Hz while the knee-point has an f$_{action}$ of 100~Hz. From point A to knee-point denotes a 100$\times$ improvement in action throughput and translates to an increase in velocity from 10 m/s to 30 m/s. However, after the knee-point, even 100$\times$ improvement in f$_{action}$ results in only 1.0004$\times$ improvement in velocity. Hence, increasing the action throughput (e.g., faster sensor or compute, etc.) beyond the knee-point yields no improvement in V$_{safe}$.  

\subsection{F-1 Model Captures the Effects of Payload Weight}

The F-1 model can visualize the effect of varying payloads such as onboard computers, sensors, and batteries. Usually, an onboard computing platform with higher TDP (thermal design power) weighs heavier due to the larger heatsink and board-level components. Payload weight such as onboard computer and larger heatsink affects UAV's acceleration (a$_{max}$). On the one hand, a larger payload weight lowers a$_{max}$, which lowers the safe velocity. The ceilings in the F-1 model capture this effect (See a$_{1}$ in  \Fig{fig:eff-a}). On the other hand, a lighter payload weight can affect the a$_{max}$ less and can allow the UAV to have a higher safe velocity. This effect is captured by the roofline or increased ceilings (a$_2$ or a$_{3}$ in \Fig{fig:eff-a}). Hence, the F-1 model helps computer architects understand how the onboard compute power, its TDP (or heatsink) impacts a$_{max}$ and safe velocity.

\section{Experimental Validation of the F-1 Model}
\label{sec:validation}

This section discusses the F-1 model validation based on the real-world UAV flight. We demonstrate that the knee-point for different drones, which determines the roof, matches the values predicted by the F-1 model within acceptable error bounds.  

\begin{figure}[t]
\vspace{5pt}
        \includegraphics[height=0.4\columnwidth,width=\columnwidth,keepaspectratio]{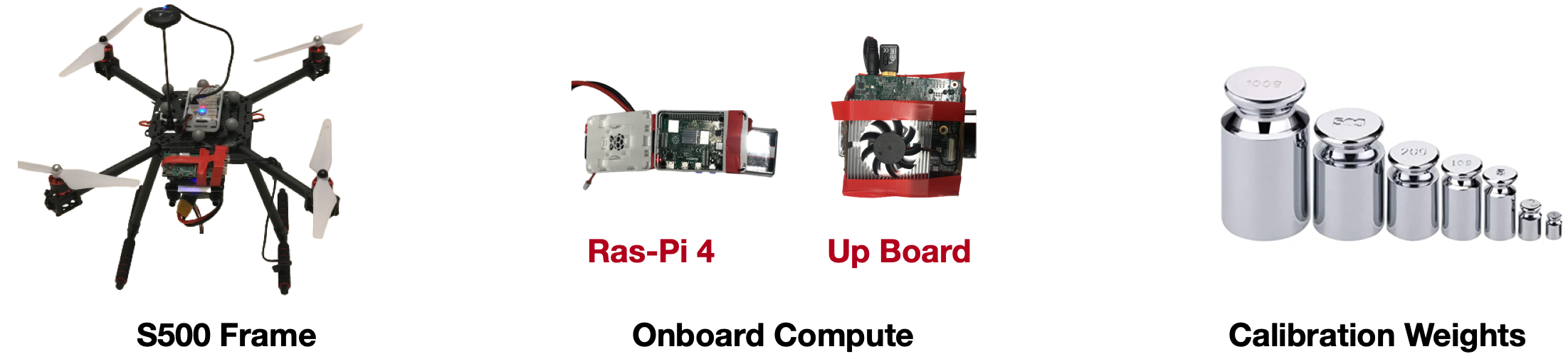}
        \vspace{-10pt}
        \caption{We build four custom drones by changing the onboard compute (Ras-Pi4 and UpBoard) and calibration weights (50 g and 150 g). The mechanical frame is an S500 quadcopter frame for all four UAVs. Note that the figure is not to scale.} 
        \label{fig:drone-builds}
\end{figure}

We built four drones by changing two onboard compute platforms (which change the payload weight) and calibration weights (to simulate different heatsink weights) while keeping the software stack and mechanical frame the same (\Fig{fig:drone-builds}). 
Our choice of the four UAV configurations was primarily based on the limitations of current UAV technology. We needed the ability to customize the UAV and have programmable control. We also needed a minimum onboard computing capability to run autonomy algorithms and several ROS packages (e.g., ground truth localization). In addition, we needed support for customization in terms of payload. Hence, we built the custom UAV from scratch using off-the-shelf components to validate Eq~\ref{eq:vmax}, as well as expand it as the F-1 roofline model to allow characterization and bottleneck analysis for other types of autonomous UAVs. In contrast, commercial off-the-shelf UAVs use proprietary components, have limited payload choice, and have almost no programmable access to the control software.

 \textbf{UAV(s) Specification.} All component specifications used in four UAVs are tabulated in Table~\ref{tab:drone-spec}. UAV-A and UAV-B use the same motor with a pull (thrust) of 435 \texttt{g} per motor. Likewise, UAV-C and UAV-D use additional calibration weights of 50~\texttt{g} and 150 \texttt{g}, which will change the thrust-to-weight ratio of the UAV. These weights simulate additional payload components. The onboard compute on UAV-A, UAV-C, and UAV-D is Ras-Pi4 (ARM-based system), whereas, on UAV-B, it is Up Squared (x86 based system). 
  Considering the entire payload system (i.e., onboard compute + battery) has a weight difference of 210 \texttt{g}, these differences result in UAVs having completely different flight characteristics. Also, due to these differences, the four drones will have unique rooflines (since each has different a$_{max}$) captured by our F-1 model. Later on, we describe how our validation methodology generalizes to other UAV types, such as nano- vs. micro-UAV, and so forth.

\begin{table}[t]
\resizebox{\columnwidth}{!}{
\centering
\begin{tabular}{|l|l|l|l|l|}
\hline
\textbf{UAV Components} & \multicolumn{1}{c|}{\textbf{\begin{tabular}[c]{@{}c@{}}UAV\\ A\end{tabular}}} & \multicolumn{1}{c|}{\textbf{\begin{tabular}[c]{@{}c@{}}UAV\\ B\end{tabular}}} & \multicolumn{1}{c|}{\textbf{\begin{tabular}[c]{@{}c@{}}UAV\\ C\end{tabular}}} & \multicolumn{1}{c|}{\textbf{\begin{tabular}[c]{@{}c@{}}UAV\\ D\end{tabular}}} \\ \hline \hline
\textit{Flight Controller} & \multicolumn{4}{c|}{NXP FMUk66} \\ \hline
\begin{tabular}[c]{@{}l@{}}Base Weight\\ (Motors + ESC + Frame)\end{tabular} & \multicolumn{1}{l|}{1030 g} & \multicolumn{1}{l|}{1030 g} & \multicolumn{1}{l|}{1030 g} & 1030 g \\ \hline
\textit{Battery} & \multicolumn{4}{c|}{3S 5000 mAh, 11.1 V} \\ \hline
\textit{Autonomy Algorithm} & \multicolumn{4}{l|}{Custom Controller based on MAVROS} \\ \hline
\textit{Onboard Compute} & \multicolumn{1}{l|}{Ras-Pi4} & \multicolumn{1}{l|}{UpBoard} & \multicolumn{1}{l|}{Ras-Pi4} & Ras-Pi4 \\ \hline
\textit{Motor Propulsion} & \multicolumn{4}{c|}{ReadytoSky 2210 920 KV} \\ \hline
Motor Pull (single motor) & \multicolumn{1}{l|}{$\approx$ 435 g} & \multicolumn{1}{l|}{$\approx$ 435 g} & \multicolumn{1}{l|}{$\approx$ 435 g} & $\approx$ 435 g \\ \hline
\begin{tabular}[c]{@{}l@{}}Payload Weight\\ (Batteries + Onboard Compute)\end{tabular} & \multicolumn{1}{l|}{590 g} &\multicolumn{1}{l|}{800 g}  & \multicolumn{1}{l|}{640 g} & \multicolumn{1}{l|}{690 g} \\ \hline
\end{tabular}}
\caption{Specification of four custom UAVs we build for F-1 model validation. Each of the four UAVs will have different real-world flight characteristics due to different takeoff weights.}
\label{tab:drone-spec}
\end{table}

\textbf{Experimental Validation Methodology.} To experimentally validate that the safe velocity (knee-point) predicted by the F-1 model matches the actual drone safe velocity, we write our custom autonomy algorithm to precisely control the drone's position, velocity, and acceleration. It is developed using the MAVROS UAV drivers running on Robot Operating System (ROS). The entire algorithm is executed on Ras-Pi4 or Up Squared. The ground truth of the UAVs' positions is captured from a well-calibrated Vicon motion capture system to verify the infractions in the stopping position. The drone's velocity and acceleration are obtained from the flight controller's IMU.

In our experiments, we start with an obstacle placed at 3 \texttt{m} from the drone's current position, and the goal of the autonomy algorithm is to move and safely stop before the obstacle. We also assume that the sensing distance is at least 3m (i.e., d = 3~\texttt{m}) to detect the obstacle and make the decision to stop. This is a reasonable setup since we can assume that the 3 \texttt{m} stop is an intermediate goal in a larger trajectory~\cite{prm-rl} which could be generated using either SPA or E2E learning (see Section~\ref{sec:algos}) algorithm. If infractions exist beyond the 3 \texttt{m}, it signifies that the drone has collided with the obstacle. Likewise, if the drone stops before the 3 \texttt{m}, it suggests that it has not collided.

For each UAV (UAV-A to UAV-D) and its specification (see Table~\ref{tab:drone-spec}), we first obtain the safe velocity predicted by the F-1 model. Using these values as the seed, we vary the drone's velocity in the seed value neighborhood. For example, the safe velocity predicted by the F-1 model for UAV-A is 2.13 \texttt{m/s}. In flight tests, we control UAV-A's velocity from 1.5 \texttt{m/s}, 1.9 \texttt{m/s}, 2 \texttt{m/s}, 2.1 \texttt{m/s}, 2.2 \texttt{m/s}, and 2.5 {m/s} respectively. 

\begin{figure}[t!]
\centering
        \begin{subfigure}{0.48\linewidth}
        \includegraphics[width=0.9\columnwidth, keepaspectratio]{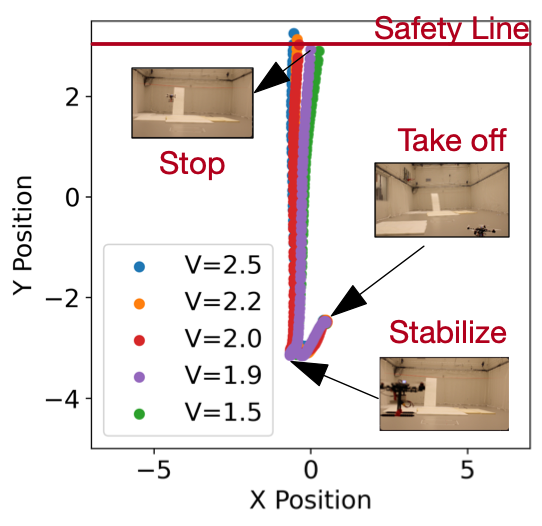}
        \caption{\footnotesize UAV-A flight trajectory.}
        \label{fig:drone_a}
        \end{subfigure}
        \begin{subfigure}{0.48\linewidth}
        \includegraphics[width=0.9\columnwidth, keepaspectratio]{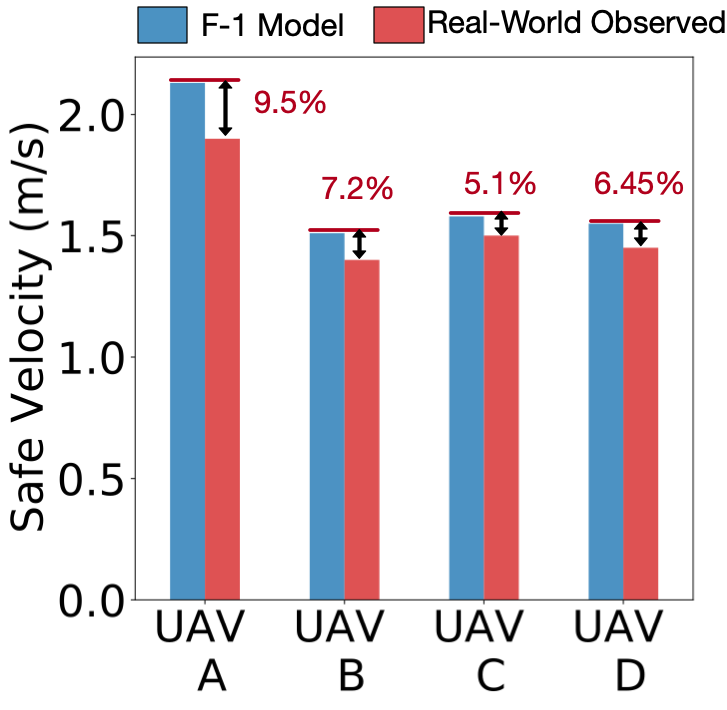}
        \caption{\footnotesize Error.}
        \label{fig:f1_error}
        \end{subfigure}
    \label{fig:fund-relation}
 \caption{(A) Real-world flight trajectory for UAV-A. The F-1 model predicted safe velocity is 2.2 \texttt{m/s}. Based on the real-world flight test, the safe velocity for UAV-A is 1.9 \texttt{m/s}. (B) Error \% between F-1 predicted and observed safe velocity.\vspace{-10pt}}
\end{figure}

\Fig{fig:drone_a} shows the trajectories obtained from UAV-A flight by setting the velocity from 1.5 \texttt{m/s} to 2.5 \texttt{m/s}. 
For the 1.5 \texttt{m/s} the UAV-A will always stop safely. For 2.5 \texttt{m/s}, the UAV-A will always have infractions. We vary the UAV-A velocity to the point where we see no infractions (v=1.9 \texttt{m/s}), corresponding to the safe velocity of UAV-A in the real world.

We run five different trails for each UAV and each velocity point to rule out the statistical noise. For instance, with 2 \texttt{m/s}, the UAV-A had infractions twice out of five trials. But we still consider this velocity to be unsafe. Upon finding the safe velocity, we calculate the error between the F-1 model's safe velocity and the real-world UAV's safe velocity for flying.

After we validate the safe velocity, we also need to validate the knee-point. Recall that the $y$-value of the knee-point is safe velocity, and the $x$-value is the action throughput that achieves this safe velocity. Since our autonomy algorithm is programmed in MAVROS, the ROS loop rate parameter sets the action throughput. Our experiments set the loop rate to 10 Hz, which matches the knee-point determined by the F-1 model for these drones. \Fig{fig:f1_error} shows the error between the F-1 model predicted safe velocity and the real-world flight. The end-to-end modeling error is 9.5\%, 7.2\%, 5.1\%, and 6.45\% for each drone type, i.e., UAV-A through UAV-D, respectively. 

\begin{figure}[t!]
\begin{minipage}{0.45\columnwidth}
\includegraphics[width=0.6\linewidth]{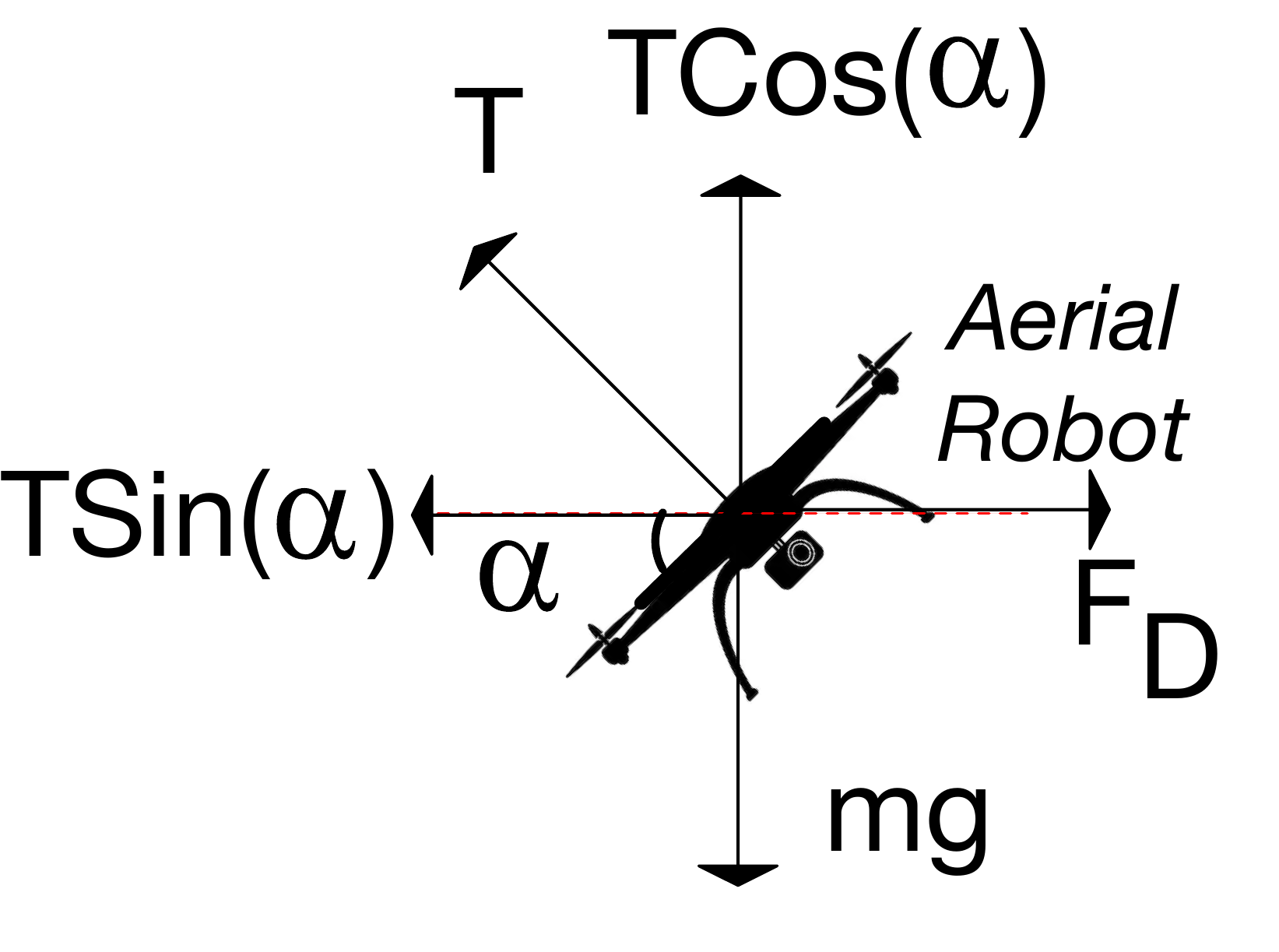}
\end{minipage}
\begin{minipage}{0.45\textwidth}
\scriptsize
\begin{equation}
\begin{split}
T\cos(\alpha) - mg = m\text{a}_{y} \\
T\sin{\alpha} - F_{D} = m\text{a}_{x} \\
\overrightarrow{{a}_{y}} = \frac{T\cos(\alpha) - mg}{m}\\
\overrightarrow{{a}_{x}} = \frac{T\sin(\alpha) - F_{D}}{m}\\
\overrightarrow{{a}_{max}} = \overrightarrow{a_{x}} + \overrightarrow{a_{y}}
\end{split}
\label{eq:acc-correction}
\end{equation}
\null
\par\xdef\tpd{\the\prevdepth}
\end{minipage}
\centering\captionof{figure}{Estimation of acceleration.}
\label{fig:fbd}
\end{figure}

\begin{figure}[t]
        \includegraphics[width=0.7\columnwidth]{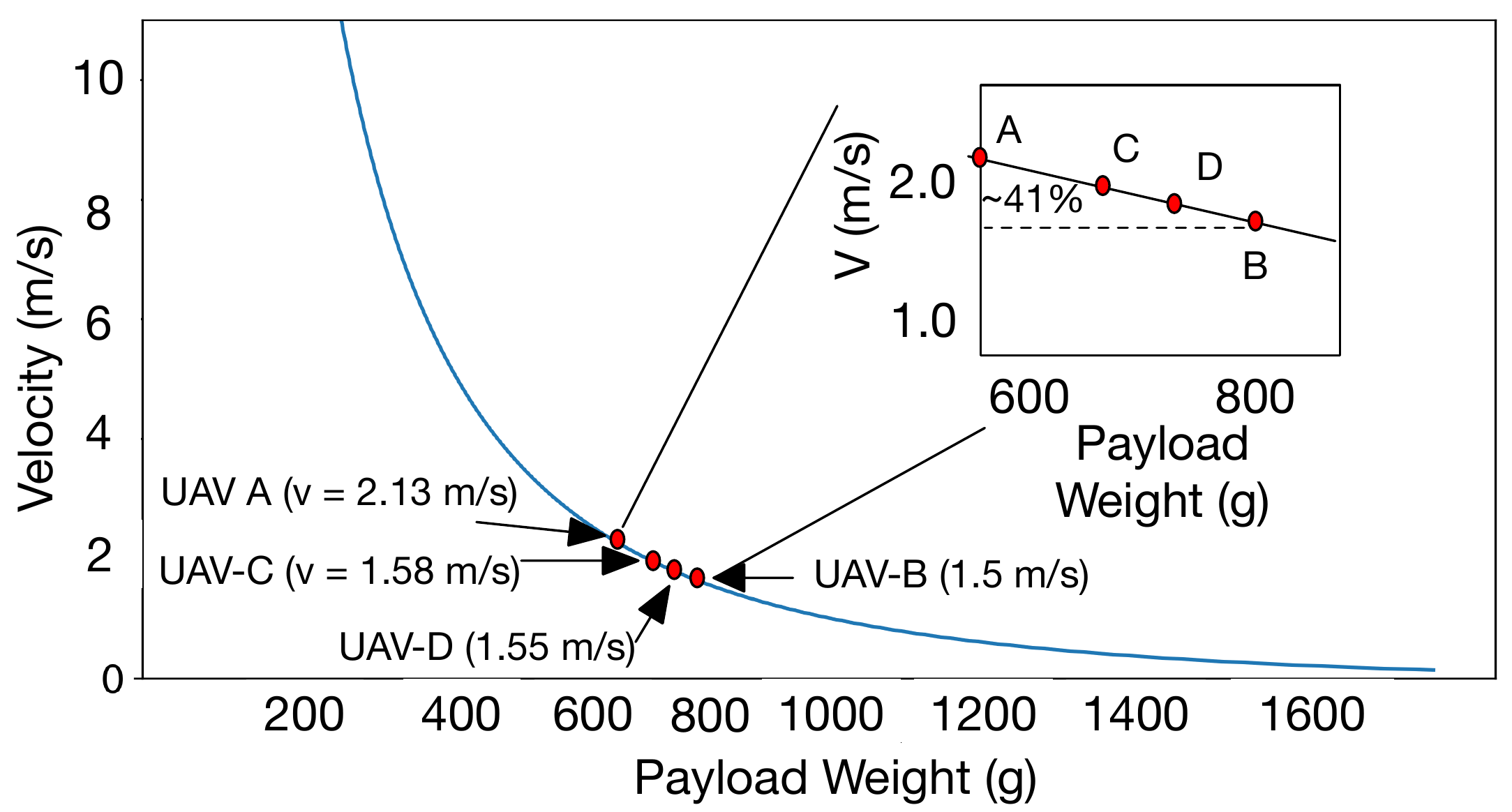}
        \caption{Non-linear relationship between the safe velocity (V$_{safe}$) and the maximum payload weight.}
        \label{fig:vel-payload-weight}
\end{figure}

Note that even though the payload weight difference (as shown in Table~\ref{tab:drone-spec}) between UAV-A and UAV-C versus UAV-C to UAV-D is the same (i.e., 50 g), the drop in safe velocity (V$_{safe}$) is not proportional in \Fig{fig:f1_error}. This is due to the non-linear relationship between the safe velocity, acceleration (a$_{max}$), and the payload weight. To determine the relationship between the velocity and payload weight, we must start from Eq~\ref{eq:vmax} and calculate the acceleration as a function of the payload weight. We estimate the upper bound in acceleration for each payload weight by using Eq~\ref{eq:acc-correction} in \Fig{fig:fbd}, where $T$ is the total thrust, $\alpha$ is the angle pitch angle, and $m$ is the mass of the payload, and \textit{F$_D$} is the drag force, which depends on the aerodynamic properties of the UAV. Since we aim to provide an early characterization and bottleneck analysis tool to guide onboard computing system design or selection, we do not model drag in the F-1 model. The F-1 model calculates the a$_{max}$ based on the payload weight. \Fig{fig:vel-payload-weight} shows V$_{safe}$ velocity as a function of the payload weight. We map the four UAV configurations in Table~\ref{tab:drone-spec} onto the velocity versus payload weight curve in \Fig{fig:vel-payload-weight}. A 50~g payload weight increase in UAV-A to UAV-C causes a $\sim$35\% decrease in velocity (from 2.13 m/s to 1.58 m/s), whereas the same 50~g weight increase from UAV-C to UAV-D results in $<$3\% reduction in velocity (1.58 m/s to 1.53 m/s). Likewise UAV-B (Intel UpBoard), which is 210 g heavier than UAV-A (Ras-Pi4), causes a $\sim$41\% decrease in safe velocity from 2.13~m/s to 1.51~m/s as shown in zoomed-in sub-plot in \Fig{fig:vel-payload-weight}. The $y$-axis in the zoomed-in subplot is in the log scale. Hence, it is important to understand which region the UAV is operating in and how the design/selection of various payloads (e.g., onboard compute) impacts UAV's performance. Generally speaking, as the UAVs get smaller in form factor, their payload carrying capability decreases, and as such, their safe velocity is more dramatically affected due to the non-linear relationship shown in \Fig{fig:vel-payload-weight}.

\textbf{Generalization to Other UAVs Beyond Table~\ref{tab:drone-spec}}. We discussed earlier that our UAV selection(s) was largely determined by the minimum capabilities that we required for the experimental validation of the F-1 model. The lowest-end onboard compute platform that can run MAVROS and other software packages is the Ras-Pi4. However, this computing platform requires a separate onboard battery (due to the limitations in UAV power delivery), weighing 590~g. We had also selected another Intel UpBoard capable of running MAVROS and autonomy algorithm as another onboard compute platform. Similar to Ras-Pi4, the UpBoard computer board also requires a separate battery. The Intel UpBoard onboard computer and battery for its power supply weigh around 800g. Hence, between these two extremes (UAV-A and UAV-B), we added calibration weights to create two additional UAV configurations (UAV-C and UAV-D) to validate the F-1 model. 

But in general, one can pick any point in the curve (signifying a different UAV configuration) from \Fig{fig:vel-payload-weight}. In the future, it would be feasible to integrate specific board-level components within the SoC with future technological advancements to reduce the payload weight further. Also, a different UAV (e.g., DJI Spark) would likely have a different payload weight sensitivity based on its thrust and other UAV characteristics. Hence, it is essential to consider these effects when selecting or designing the onboard computer for UAVs. Existing efforts in designing custom hardware/miniaturization (Section~\ref{sec:discussion}) are based on isolated compute metrics without understanding how these design choices impact the UAV's flight performance. Hence, there is a need for a model like F-1 that provides abstractions similar to the Roofline model~\cite{roofline} to allow computer architects to characterize onboard computers and determine various bottlenecks affecting UAV's performance.

\textbf{Understanding the Potential Source(s) of Error(s).} Although we envision the F-1 model to be used as an early-phase design tool to characterize the computing system for UAVs, it is crucial to understand the source of errors to compensate for it during the later part of the design cycle. We believe there are three reasons for this error. First, the F-1 model introduces a linearization error when converting the curve to the left of the knee-point with a straight line (\Fig{fig:roofline-drone-logx}). Second, the  F-1 model calculates the impact of computing weight
on a drone's acceleration. However, in real-world flight, drag (\Fig{fig:fbd})
can also impact the drone's acceleration. Third, we build the drone from the ground up (sourcing various components) and attach payload weights directly to the frame. The sudden movements (e.g., jerk) of the payload components can affect the drone's dynamics and center of mass. We believe that with precise mechanical design, one can minimize these errors.

\textbf{Implications for Onboard Computer Design.} The F-1 model predicts a higher safe velocity than the real-world observation. For computer architects who want to use the F-1 model to understand different bottlenecks, the optimistic estimation of the F-1 model is beneficial. Any onboard compute designed at the v$_{safe}$ predicted by F-1 will also ensure that the design will not affect the a$_{max}$ at a lower safe velocity. Hence, overestimation by the F-1 model will always ensure that the onboard compute will never be the bottleneck.

\textbf{Summary.} The experimental validation results show that the F-1 model as a visual performance model is, on average, 90\% to 95\%  accurate in predicting the knee-point (roofline) of the drone just entirely based on its specification. We now extend the validated F-1 model to construct a web-based tool (called Skyline) so that system  architects can use this interactive tool to easily characterize the computing system or use it to design domain-specific hardware accelerators for autonomous drones.

\begin{table}[]
\normalsize
\renewcommand\arraystretch{1}
\resizebox{0.95\textwidth}{!}{
\begin{tabular}{|c|c|l|}
\hline
\textbf{\begin{tabular}[c]{@{}l@{}}Parameter\end{tabular}} & \textbf{Unit} &  \textbf{Description} \\ \hline
\textit{\begin{tabular}[c]{@{}c@{}}Sensor \\ Framerate\end{tabular}} & \texttt{Hz}  &  Throughput of the sensor. \\ \hline
\textit{\begin{tabular}[c]{@{}c@{}}Compute \\ TDP\end{tabular}} & \texttt{W} &  \begin{tabular}[c]{@{}l@{}}Maximum TDP of the onboard\\ compute. Used to design the heatsink.\end{tabular} \\ \hline
\textit{\begin{tabular}[c]{@{}l@{}}Autonomy\\ Algorithm\end{tabular}} & N/A &  \begin{tabular}[c]{@{}l@{}}Select a pre-configured autonomy algorithm.\end{tabular} \\ \hline
\textit{\begin{tabular}[c]{@{}l@{}}Compute\\ Runtime\end{tabular}} & \texttt{s} &  \begin{tabular}[c]{@{}l@{}}Measures the latency of the autonomy\\ algorithm. Used to calculate compute throughput.\end{tabular} \\ \hline
\textit{\begin{tabular}[c]{@{}c@{}}Sensor \\ Range\end{tabular}} & \texttt{m} &  Maximum range of the sensor. \\ \hline
\textit{\begin{tabular}[c]{@{}l@{}}Drone \\ Weight\end{tabular}} & \texttt{g} &  \begin{tabular}[c]{@{}l@{}}Maximum weight of the UAV without\\any extra payload.\end{tabular} \\ \hline
\textit{\begin{tabular}[c]{@{}l@{}}Rotor \\ Pull\end{tabular}} & \texttt{g} &  \begin{tabular}[c]{@{}l@{}}Measures the thrust produced by the \\ rotor propulsion.\end{tabular} \\ \hline
\textit{\begin{tabular}[c]{@{}l@{}}Payload\\ Weight\end{tabular}} & \texttt{g}&  \begin{tabular}[c]{@{}l@{}}Total weight of the payload including\\ onboard compute, sensors, battery etc.\end{tabular} \\ \hline
\end{tabular}}
\caption{Summary of knobs available in the Skyline tool.}
\label{tab:knobs-table}
\end{table}

\section{Skyline: An Interactive Visual Tool for F-1}

\begin{figure*}[t]
\centering
\begin{subfigure}{\textwidth}
\centering
        \includegraphics[trim=385 0 0 50, clip, width=\textwidth]{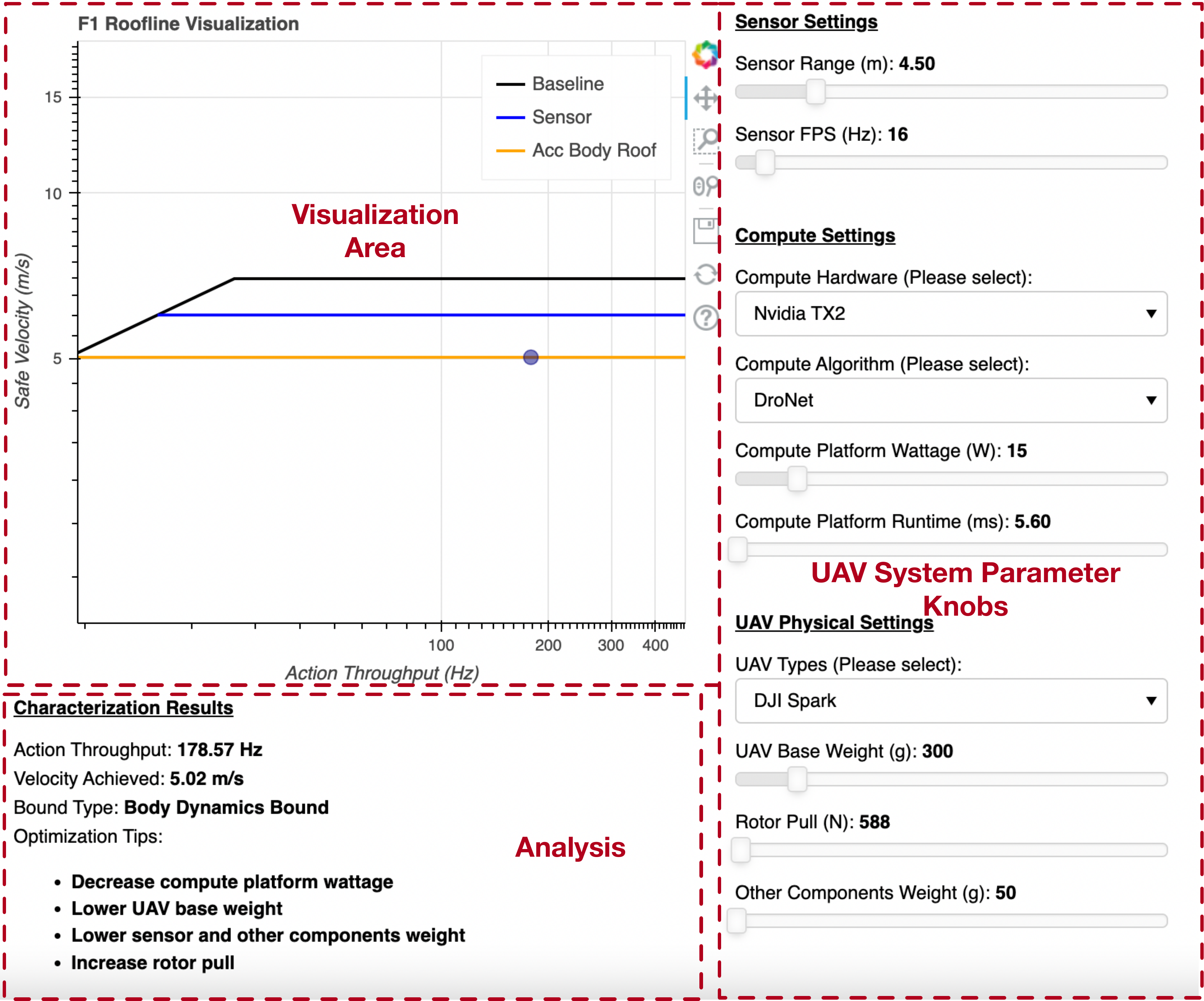}
        
\end{subfigure}%
    
 \caption{Skyline is an interactive tool to visualize the F-1 model in action. It allow users to change various UAV and compute system parameters and observe the resulting end-to-end effects. The tool also provides recommendations for easy data analysis.
 \vspace{-0.2in}}
 \label{fig:tool}
\end{figure*}

We provide an interactive web-based tool (Figure~\ref{fig:tool}) to plot the F-1 model for a given UAV. It gives the end-user the ability to do exploratory studies on the impact of various parameters of the UAVs, such as algorithms, onboard computers, and sensors.

\subsection{Overview}

The tool has three major components. The first component includes the interactive knobs for various UAV components and their parameters. The second component is the visualization area that dynamically plots the F-1 model. The third component is the analysis and guidance area, where the tool shows the characterization results and optimization tips for the architects. 

\subsection{UAV System Parameter Knobs}

We divide the parameters into three categories: sensor, compute, and physics. Recall that the F-1 model (Section~\ref{sec:f-1-roofline}) ties the various UAV component interactions together to determine the safe velocity. The maximum safe velocity has implications on the overall mission performance~\cite{mavbench}. Table~\ref{tab:knobs-table} summarizes the available knobs in the Skyline tool. 

The tool, as shown in Figure~\ref{fig:tool}, also provides user-defined knobs. These knobs all the end-user to specify unique parameter values for sensor range, sensor framerate, compute platform TDP, compute latency, UAV weight, and rotor thrust. These parameters can be used for future design characterization studies to understand how these parameters impact the UAV's performance. Later on, in the results section, we demonstrate how to use these knobs via a variety of different case studies.

\subsection{Visualization Area}
The visualization area plots the F-1 model based on the UAV configurations. The interactive UAV parameter knobs result in an interactive plot that gives the user an understanding of how these parameters affect the UAV's safe velocity. For example, increasing the TDP of the compute platform increases the heatsink and payload weight. As payload weight increases, the maximum acceleration decreases, lowering the safe velocity. Likewise, changing the sensor framerate affects the decision-making rate, determining how fast the UAV can travel. Thus, using the interactive visualization of the F-1 model enables the user to intuitively understand the effect of UAV parameters.

\subsection{Automatic Analysis}
The Skyline tool has an analysis section that outputs information, such as knee-point throughput and safe velocity achievable, for a given UAV. 
In addition to the performance data, the analysis section also provides information about the fundamental limits in safe velocity (i.e., sensor bound, compute-bound, or physics bound) and several optimization tips.

\section{Evaluation}
\label{sec:eval}

We present several case studies (Table~\ref{tab:overview}) on how to use the F-1 model for the characterization of various components in a UAV. The analysis gathers architectural insights that can build efficient onboard compute platforms.
From Section~\ref{sec:comp} to Section~\ref{sec:dcr}, we first show how to use the model for characterizing individual components in the UAV system. Then, in Section~\ref{sec:full}, we demonstrate how the model can be used for complete end-to-end UAV system characterization. 

\subsection{Onboard Compute Characterization}
\label{sec:comp}

\begin{figure}[t!]
\centering
\begin{subfigure}{0.45\columnwidth}
\hspace{-10pt}
\centering
        \includegraphics[width=1.0\columnwidth,height=1.3in,keepaspectratio]{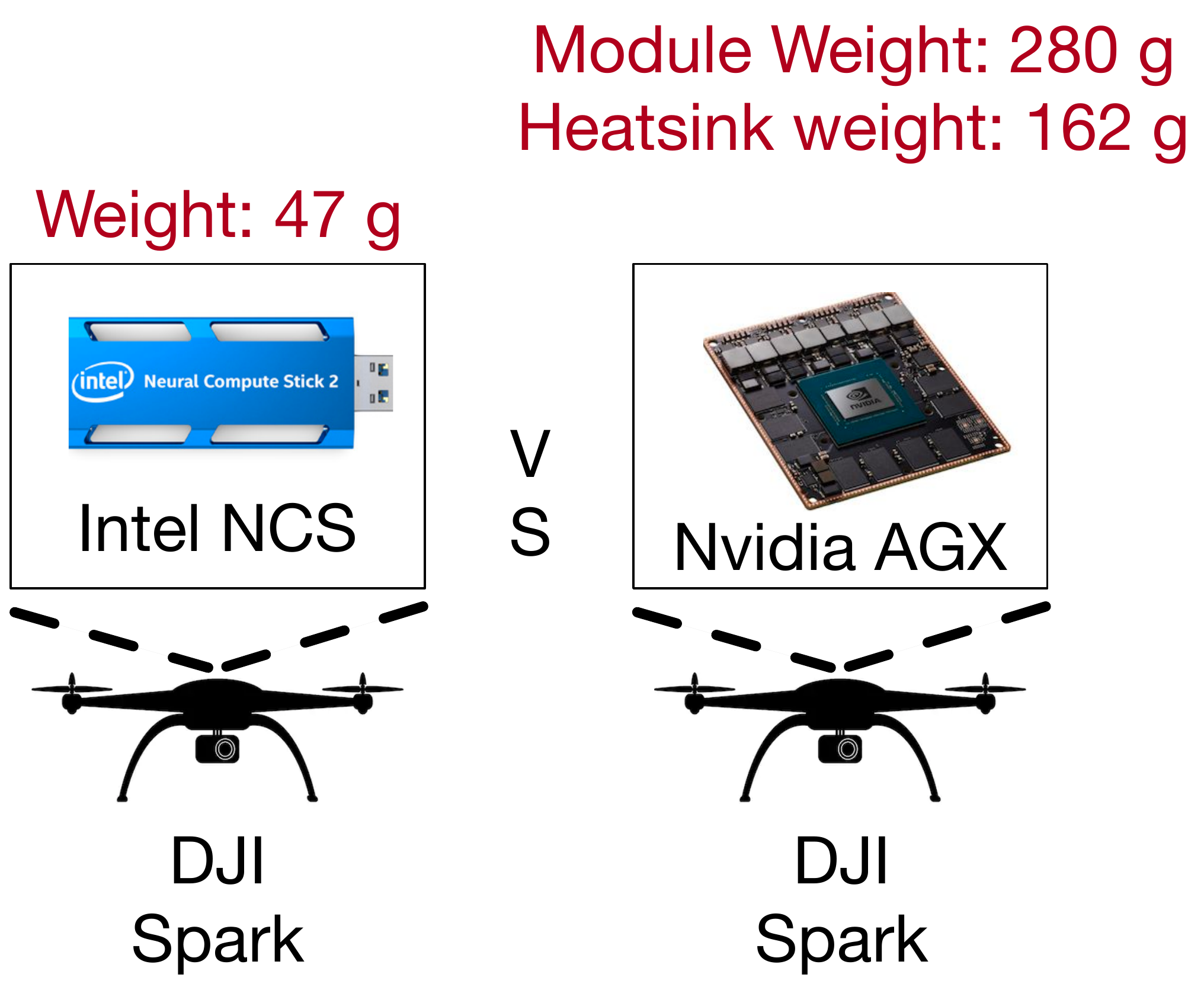}
        \caption{Different onboard compute.}
        \label{fig:diff-comp}
        \end{subfigure}
        \hspace{5pt}
        \begin{subfigure}{0.5\columnwidth}
        \includegraphics[width=1.1\columnwidth,height=1.4in, keepaspectratio]{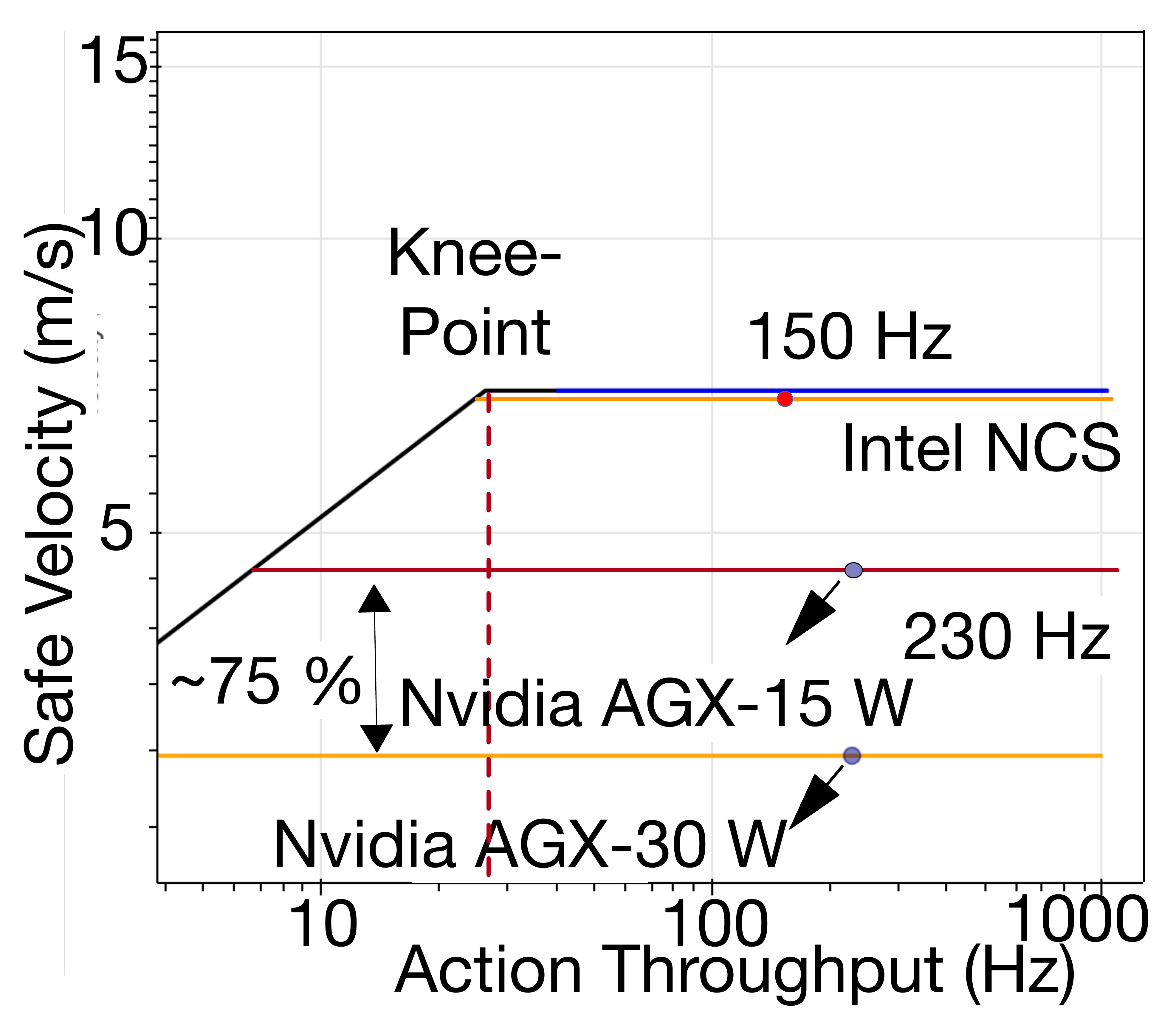}
        
        \caption{F-1 Plot.}
        \label{fig:roofline-diff-comp}
        \end{subfigure}
    \label{fig:fund-relation}
  \caption{Case study of choosing between Intel NCS and Nvidia AGX for DJI Spark running DroNet autonomy algorithm.\vspace{-10pt}} 
\end{figure}

In this case study, we answer the following research questions: \textit{Given two or more onboard compute choices, how to select a suitable computer system for a given UAV?}
The onboard computer is a vital component for achieving autonomy in UAVs. However, despite its importance, the selection (and even design, as we show in Section~\ref{sec:discussion}) of onboard compute remains ad-hoc~\cite{high-speed-drone,trailnet,krishnan2021autosoc}. We characterize two commercially available off-the-shelf onboard compute systems, Nvidia AGX and Intel NCS (shown in \Fig{fig:diff-comp}), and examine their impact.

\textbf{UAV Configurations.} For the autonomy algorithm, we select the DroNet~\cite{dronet}. Our interactive tool, Skyline, provides a few autonomy algorithms for out-of-the-box usage. We then select the DJI Spark UAV as the form factor. For the onboard compute engine, we toggle between Intel NCS and Nvidia AGX. Intel NCS (USB-like form factor) is a sub-1~W compute system that weighs around 47~g. The Nvidia AGX module without a heatsink weighs 280~g. The tool internally calculates the heatsink weight~\cite{heat-sink}, which for a 30 W TDP is 162~g. We keep the sensor FPS at 60 Hz to ensure we are not in the sensor-bound region for both these UAVs.


\textbf{Analysis.} The F-1 plot from Skyline is shown in \Fig{fig:roofline-diff-comp}. Based on the F-1 plots for the DJI Spark UAV, the Nvidia AGX (denoted as Nvidia AGX-30W) has a lower roofline than Intel NCS. The higher roofline for Intel NCS is because the NCS onboard compute weighs less than the Nvidia AGX, which means the DJI Spark with Intel NCS can achieve higher acceleration (a$_{max}$) compared to DJI Spark with Nvidia AGX. Hence, the UAV's physics restricts it from achieving a higher safe velocity between these onboard compute choices even though Nvidia AGX (230 FPS) can achieve 1.5$\times$ more compute throughput than Intel NCS (150 FPS) running the DroNet~\cite{dronet} algorithm. This suggests that the high compute performance cannot always translate to a higher safe velocity as the UAV's body dynamics/physics becomes the limiting factor.  

\begin{figure}[b]
\vspace{5pt}
        \includegraphics[width=0.5\columnwidth]{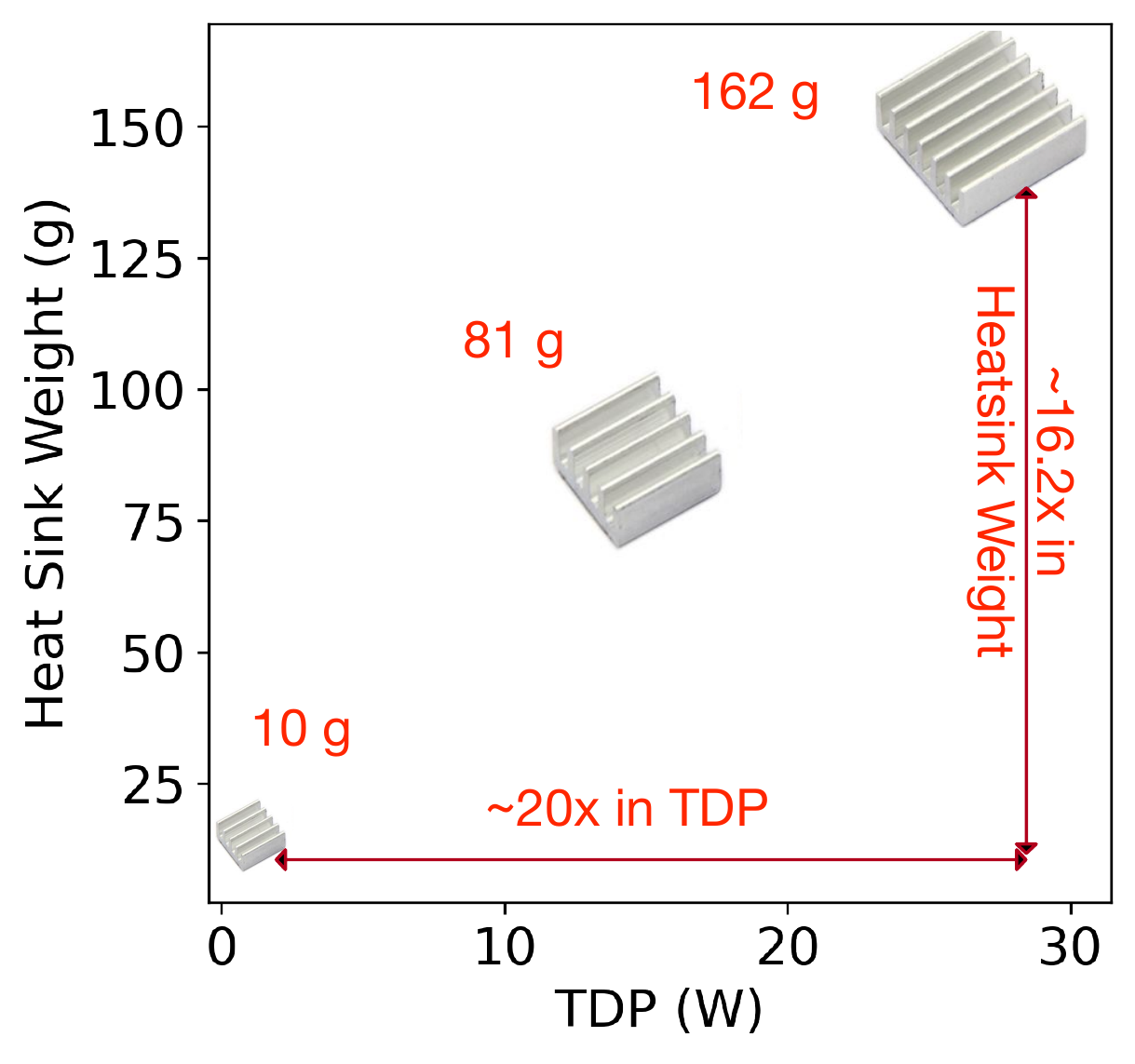}
        \vspace{-10pt}
        \caption{Reduction in heat-sink weight with TDP.}
        \label{fig:heat-sink}
\end{figure}

To improve the respective UAV's performance, a future optimization for the next iteration of Nvidia AGX could lower the power of Xavier AGX at the cost of performance (since AGX is over-provisioned by 33x). This power reduction will reduce the TDP, resulting in a smaller heatsink weight. To demonstrate the reduction of heatsink weight, we consider a scenario where we reduce the TDP of AGX from 30 W to 15 W using any architectural optimization. For simplicity, we assume this is achieved without impacting the compute throughput. Based on the heatsink calculator~\cite{heat-sink} used in the F-1 model, we observe that the reduction in TDP also reduces the heatsink weight by half (from 162 g to 81 g as shown in \Fig{fig:heat-sink}). Furthermore, the reduction of the compute payload weight increases the DJI Spark's safe velocity by 75\% as shown by the ceiling (denoted as Nvidia AGX (15W)) in \Fig{fig:roofline-diff-comp}.

\begin{table}[]
\renewcommand\arraystretch{1.1}
\resizebox{\textwidth}{!}{
\begin{tabular}{|c|c|cccc}
\hline
 &  & \multicolumn{4}{c|}{\textbf{UAV System Configurations}} \\ \cline{3-6} 
\multirow{-2}{*}{\textbf{\begin{tabular}[c]{@{}c@{}}Case\\ Studies\end{tabular}}} & \multirow{-2}{*}{\textbf{Comments}} & \multicolumn{1}{c|}{\textbf{\begin{tabular}[c]{@{}c@{}}Onboard \\ Compute\end{tabular}}} & \multicolumn{1}{c|}{\textbf{\begin{tabular}[c]{@{}c@{}}Autonomy\\ Algorithm\end{tabular}}} &  \multicolumn{1}{c|}{\textbf{\begin{tabular}[c]{@{}c@{}}Payload\\ Redudancy\end{tabular}}} & \multicolumn{1}{c|}{\textbf{\begin{tabular}[c]{@{}c@{}}UAV\\ Type\end{tabular}}} \\ \hline
\textit{VI-.A} & \textit{\begin{tabular}[c]{@{}c@{}} Onboard \\ compute\end{tabular}} & \multicolumn{1}{c|}{\cellcolor[HTML]{FFCCC9}\textit{\begin{tabular}[c]{@{}c@{}}Intel NCS\\ \&\\ Nvidia AGX\end{tabular}}} & \multicolumn{1}{c|}{\textit{DroNet}~\cite{dronet}} &  \multicolumn{1}{c|}{\textit{None}} & \multicolumn{1}{c|}{\textit{\begin{tabular}[c]{@{}c@{}}DJI \\ Spark\end{tabular}}} \\ \hline
\textit{VI-B} & \textit{\begin{tabular}[c]{@{}c@{}}Autonomy\\ algorithms\end{tabular}} & \multicolumn{1}{c|}{\textit{\begin{tabular}[c]{@{}c@{}}Nvidia\\ TX2\end{tabular}}} & \multicolumn{1}{c|}{\cellcolor[HTML]{FFCCC9}\begin{tabular}[c]{@{}c@{}}\textit{Sense-Plan-Act}~\cite{mavbench}\\ \&\\ \textit{TrailNet}~\cite{trailnet}\end{tabular}}  & \multicolumn{1}{c|}{\textit{None}} & \multicolumn{1}{c|}{\textit{\begin{tabular}[c]{@{}c@{}}AscTec\\ Pelican\end{tabular}}} \\ \hline
\textit{VI-C} & \textit{\begin{tabular}[c]{@{}c@{}}Payload\\ Redudancies\end{tabular}} & \multicolumn{1}{c|}{\textit{\begin{tabular}[c]{@{}c@{}}Two \\ Nvidia TX2\end{tabular}}} & \multicolumn{1}{c|}{\textit{DroNet}~\cite{dronet}}  & \multicolumn{1}{c|}{\cellcolor[HTML]{FFCCC9}\textit{\begin{tabular}[c]{@{}c@{}}Dual Modular \\ Redudancy\end{tabular}}} & \multicolumn{1}{c|}{\textit{\begin{tabular}[c]{@{}c@{}}AscTec\\ Pelican\end{tabular}}} \\ \hline


\textit{VI-D} & \textit{\begin{tabular}[c]{@{}c@{}}Full\\  UAV \\ System\end{tabular}} & \multicolumn{1}{c|}{\cellcolor[HTML]{FFCCC9}\textit{\begin{tabular}[c]{@{}c@{}}Nvidia TX2\\ Nvidia AGX\\ Intel NCS\\ Ras-Pi\end{tabular}}} & \multicolumn{1}{c|}{\cellcolor[HTML]{FFCCC9}\begin{tabular}[c]{@{}c@{}}\textit{CAD2RL}~\cite{cad2rl}\\ \textit{DroNet}~\cite{dronet}\\ \textit{TrailNet}~\cite{trailnet}\end{tabular}} & \multicolumn{1}{c|}{\cellcolor[HTML]{FFCCC9}\textit{None}} & \multicolumn{1}{c|}{\cellcolor[HTML]{FFCCC9}\textit{\begin{tabular}[c]{@{}c@{}}AscTec\\ Pelican\\ \&\\ DJI\\ Spark\end{tabular}}} \\ \hline
\end{tabular}}
\caption{Overview of the evaluation case study detailed in Section~\ref{sec:eval}. The highlighted parameter cell is varied while the rest of the UAV's parameters are kept constant.}
\label{tab:overview}
\end{table}

We can employ any means to optimize for performance improvement or power reduction, such as software optimizations, runtime scheduling, microarchitecture design, device-level optimization, etc. Regardless of the applied method, once we have the final performance and power (TDP) for a given autonomy algorithm, we can enter it in the Skyline tool. The tool calculates the heatsink weight based on the heatsink calculator~\cite{heat-sink} while taking into account the other UAV parameters to plot the F-1 roofline as shown in \Fig{fig:roofline-diff-comp}.


\textbf{Takeaway.} Selection of onboard cannot be ad-hoc or chosen based on compute performance in isolation, and one must consider the effects of all the UAV parameters and understand what implications the compute has on these parameters. A high-performance computer does not necessarily translate into a high-performing UAV. Skyline can help architects understand the fundamental role of computing in autonomous UAVs.

\subsection{Autonomy Algorithm Characterization}
\label{sec:algo}

In this case study, we answer the following question: \textit{Given a fixed UAV and onboard computer, what is the effect of different autonomy algorithms on V$_{safe}$?} The primary function of an autonomy algorithm is to make intelligent decisions to achieve the mission goals while guaranteeing safety. However, despite its importance, the selection or design of the algorithm is often made in isolation without considering the impact of onboard computer or UAV components. So we characterize how changing autonomy algorithms impact the UAV's performance.

\textbf{UAV Configurations.} We consider two autonomy algorithm paradigms, namely SPA and E2E  (Section~\ref{sec:background}), and evaluate them on an AscTec Pelican UAV with TX2. For SPA, we characterize the package-delivery application from MAVBench~\cite{mavbench}. For the E2E algorithm, we choose TrailNet~\cite{trail-net} and DroNet~\cite{dronet}. In both of these cases, we keep the UAV, onboard computer fixed, and evaluate the different autonomy algorithms to understand their impact on the UAV's maximum safe flying velocity. 

\begin{figure}[]
\centering
\begin{subfigure}{0.45\columnwidth}
        \includegraphics[width=0.95\columnwidth,keepaspectratio]{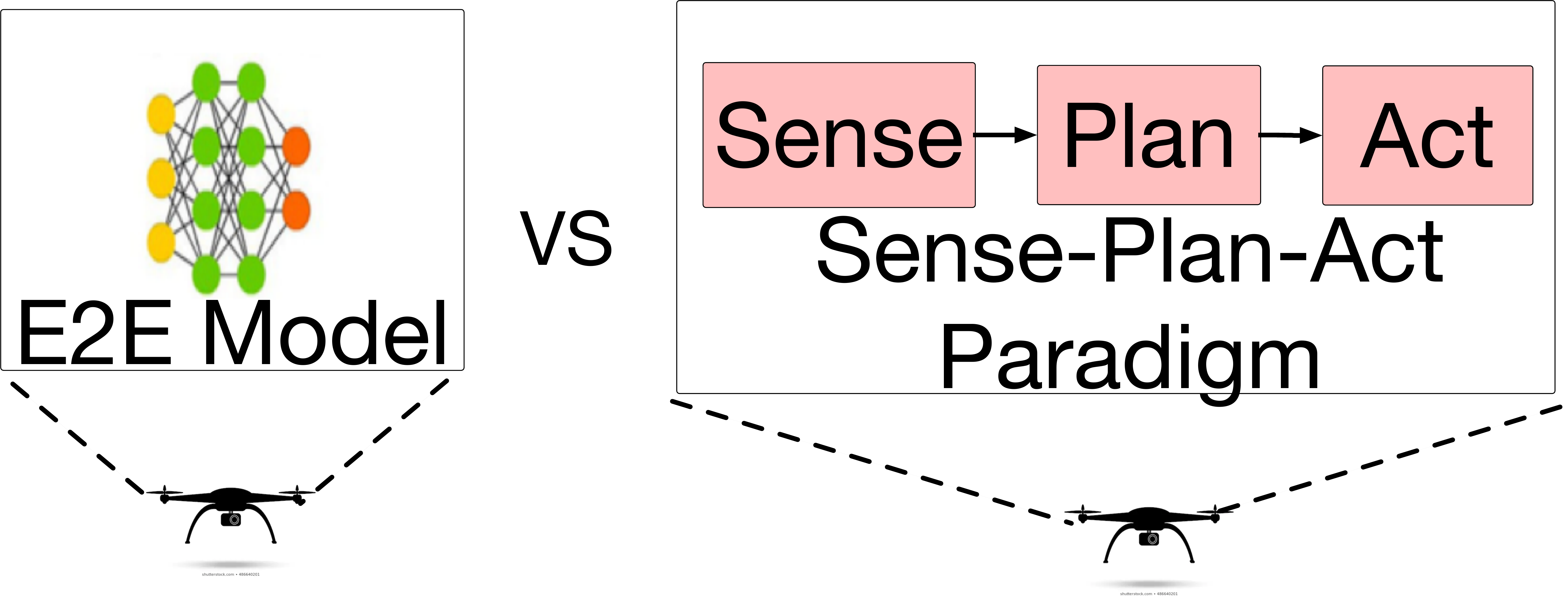}
        \caption{Different algorithms.}
        \label{fig:e2e_control}
        \end{subfigure}
        \hspace{5pt}
        \begin{subfigure}{0.45\columnwidth}
        \includegraphics[width=0.95\columnwidth, keepaspectratio]{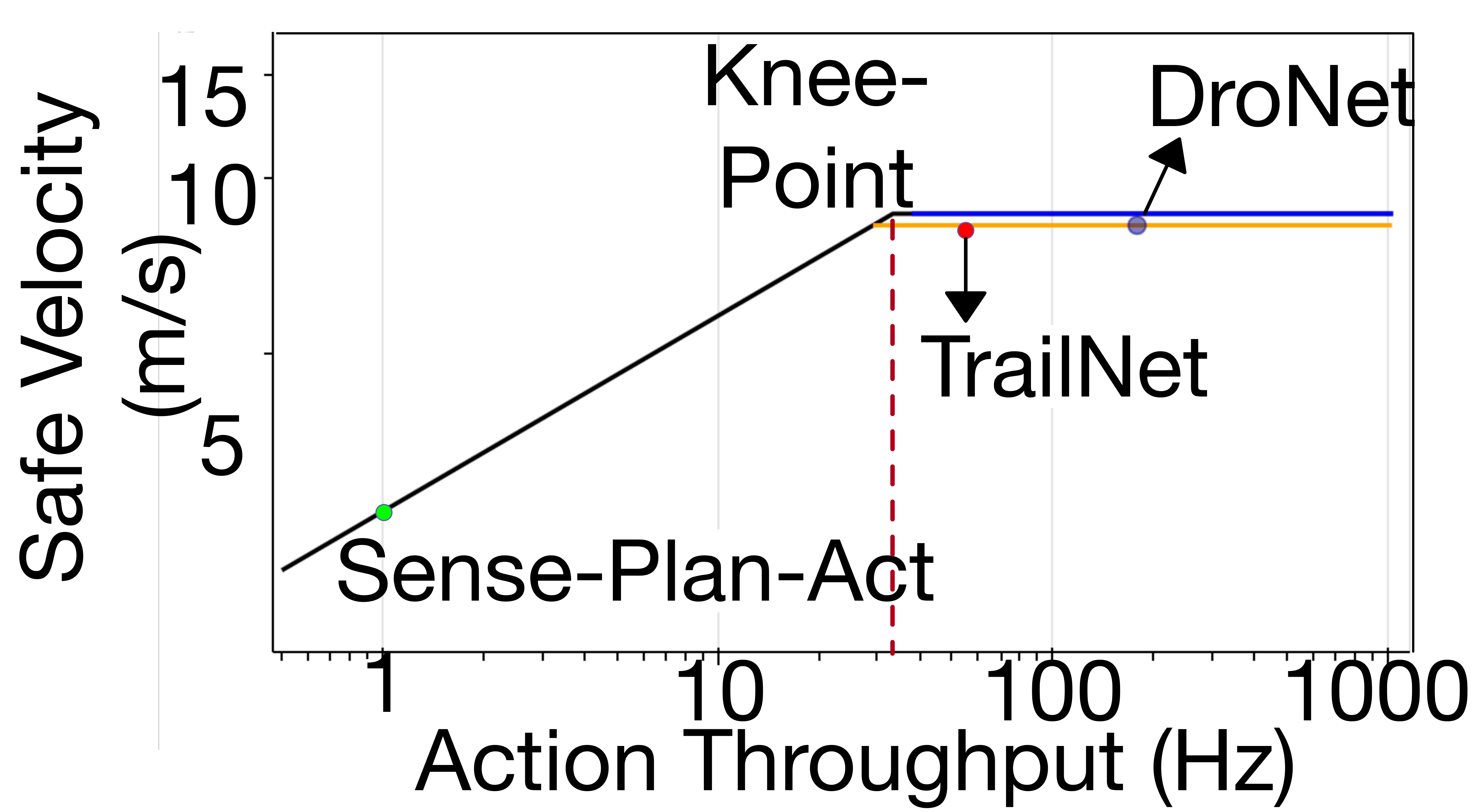}
        \caption{F-1 plot.}
        \label{fig:roofline-algos}
        \end{subfigure}
    \label{fig:fund-relation}

  \caption{Changing autonomy algorithms on the UAV. For SPA, we characterize the `package delivery' application from MAVBench~\cite{mavbench}. For E2E, we consider TrailNet~\cite{trailnet,dronet}.}
\end{figure}

\textbf{Analysis.} The resulting plots are shown in \Fig{fig:roofline-algos}. On the one hand, the SPA algorithm achieves a compute throughput of 1.1 Hz on  Nvidia TX2. However, due to its low decision-making rate, the maximum achievable safe velocity is limited to 2.3~m/s. On the other hand, TrailNet and DroNet (E2E algorithms) achieve a compute throughput of 55 Hz and 178 Hz on Nvidia TX2, which achieves a higher safe velocity.

For an AscTec Pelican UAV with the TX2 onboard compute, the knee-point throughput is 43 Hz, suggesting that TrailNet and DroNet are over-provisioned by 1.27~$\times$  and 4.13~$\times$. In the E2E paradigm, the high compute throughput of the autonomy algorithm does not translate to a higher safe velocity unless the UAV's physics changes (i.e., a UAV with a higher thrust-to-weight ratio). However, for the SPA paradigm, the UAV's safe velocity is compute-bound. Therefore, the compute throughput needs to be improved by 39 $\times$ to achieve the optimal safe-velocity permitted by the UAV's physics.

\textbf{Takeaway.} When designing autonomy algorithms for UAVs, metrics like high compute throughput or energy efficiency can be misleading and can result in over-optimization (increases the design cost) or under-optimization (affects robot performance). In an over-optimized scenario, the higher compute performance might not result in higher UAV performance. In an under-optimized scenario, the F-1 model can give us the performance targets to achieve in their optimization efforts, thereby helping us design balanced onboard computers for UAVs.

\subsection{Modular Redundancy Characterization}
\label{sec:dcr}

In this case study, we answer the following research questions: \textit{Given a UAV with fixed autonomy algorithm and sensor, what is the impact of redundancy in onboard compute on the UAV's performance?} UAVs need to operate robustly in dynamic environments~\cite{wan2021analyzing,wan2022frlfi}, and redundancy in compute or sensor ensures safety in the event of a failure. Sometimes dual redundancy~\cite{bannon2019computer} (or triple redundancy~\cite{mehmed2020monitor}) increases reliability where a majority vote determines the final decision. However, though redundancy can increase reliability, it can also increase the cost and negatively affects the UAV's performance. So we characterize the effects of adding redundancy in onboard compute and evaluate its effects on the UAV's safe velocity.

\begin{figure}[t!]
\begin{subfigure}{0.42\columnwidth}
\centering
        \includegraphics[width=1.2\columnwidth,height=1.25in,keepaspectratio]{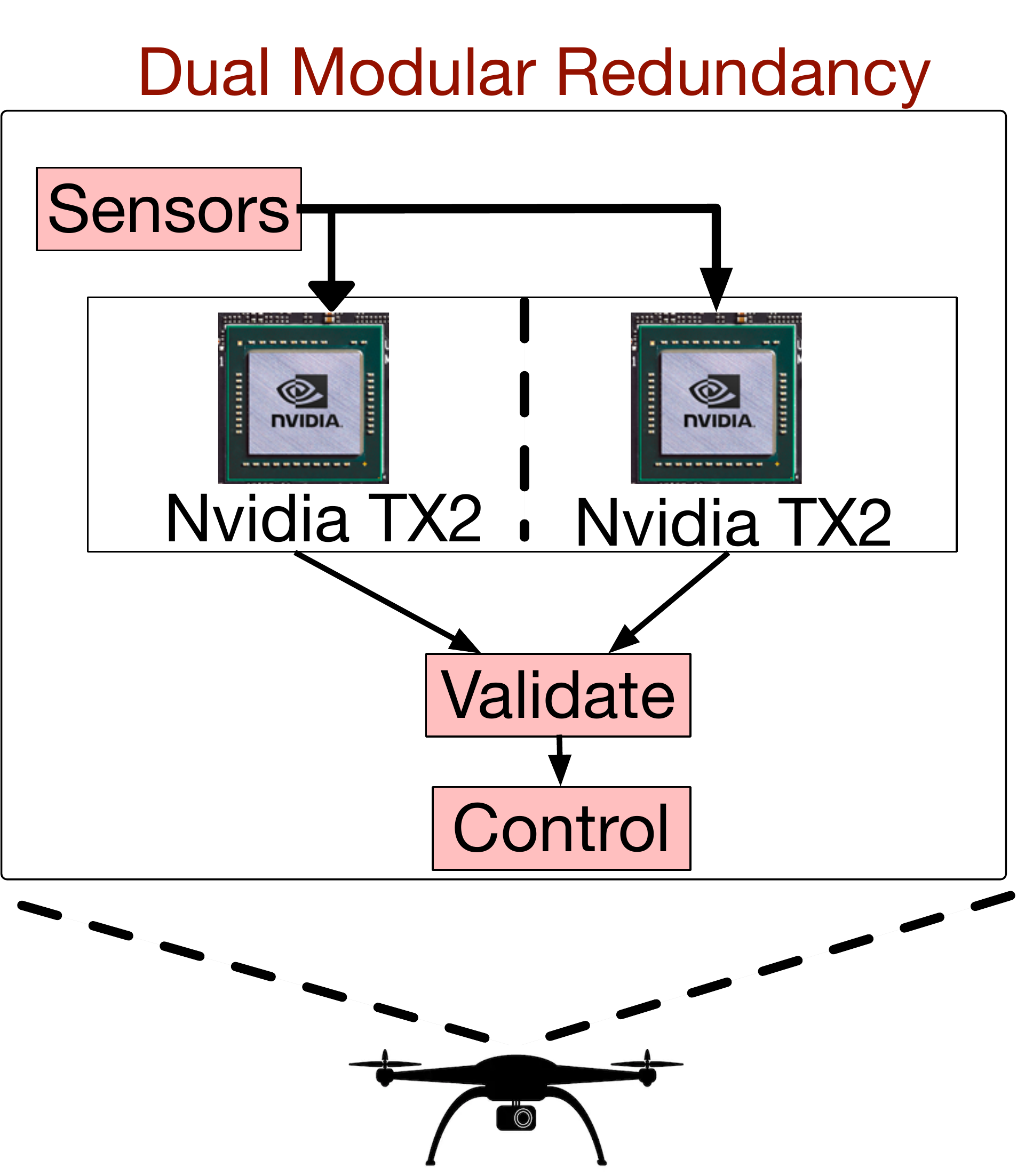}
        \caption{DMR.}
        \label{fig:dcr}
        \end{subfigure}
        \begin{subfigure}{0.55\columnwidth}
        \centering
        \includegraphics[width=0.9\columnwidth,height=1.2in, keepaspectratio]{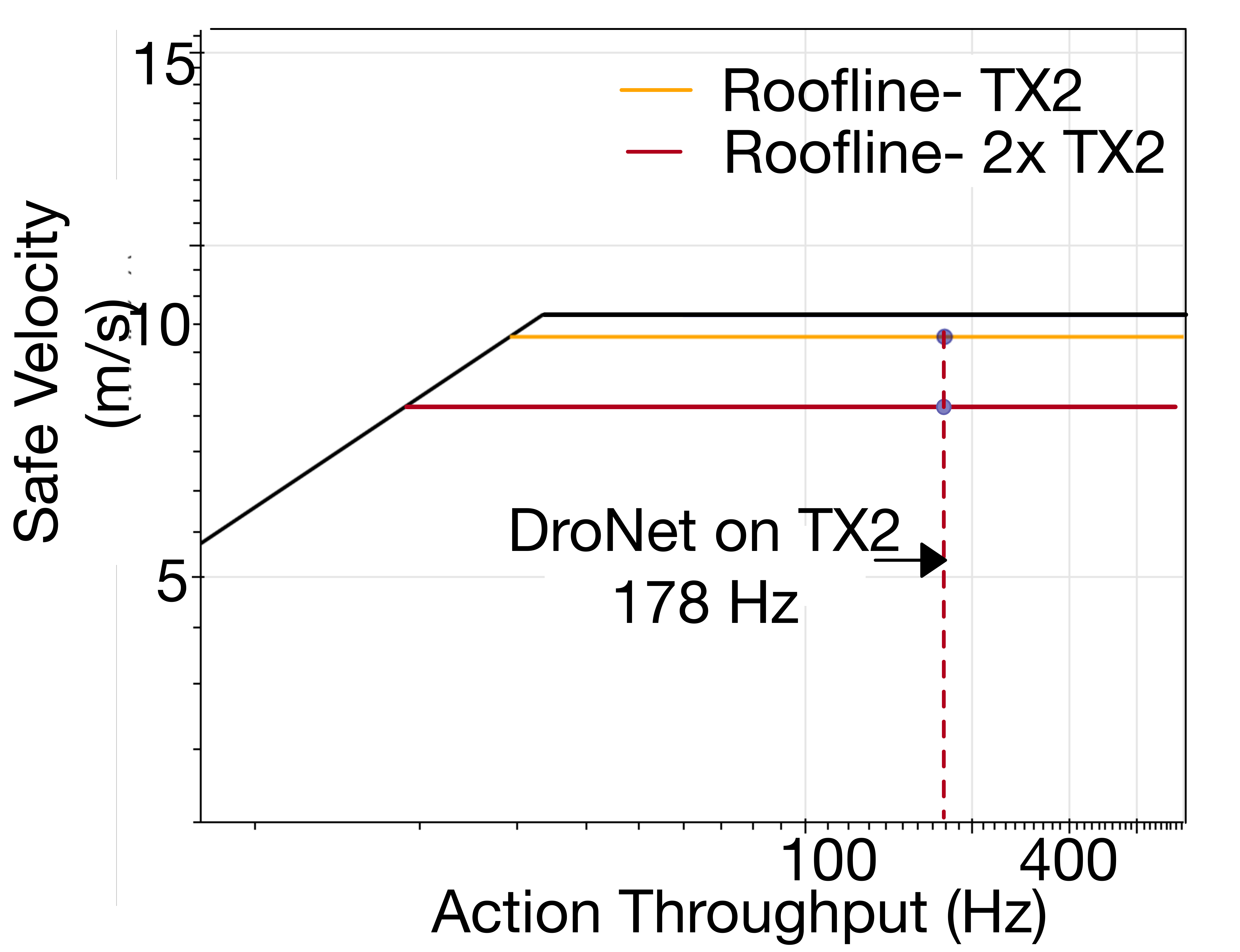}
        \caption{F-1 plot.}
        \label{fig:roofline-drc}
        \end{subfigure}
  \caption{Characterizing the effect of modular redundancy.}
\end{figure}

\textbf{UAV Configurations.} We evaluate how adding dual compute redundancy affects the UAV's safe velocity. Using  Skyline, we first select UAV~\cite{trail-net} from the pre-configured suite of autonomy algorithms. We select the AscTec Pelican UAV and an RGB-D camera with a frame rate of 60 FPS and a sensing distance of 4.5 m. In this UAV configuration, we evaluate the effects of having a single Nvidia-TX2 and dual Nvidia-TX2 SoC sharing the same board and evaluate the effects of dual compute redundancy. We select the Nvidia-TX2 platform to estimate the baseline safe velocity. To evaluate the effects of having dual redundancy in onboard compute (as shown in \Fig{fig:dcr}), we assume that another Nvidia-TX2 is added to the UAV platform and has the sensor input. The output from these two platforms is validated and then sent to the controller (similar to Tesla's FSD stack). To model this scenario using Skyline, we account for the payload weight for the additional TX2, including the computing platform and the heatsink weight.


\textbf{Analysis.} The resulting two plots are combined and shown in \Fig{fig:roofline-drc}. Since the same autonomy algorithm (and onboard compute) is used in both these UAVs, it achieves a throughput of 178~Hz running on Nvidia TX2. The baseline scenario (single onboard compute) is annotated as ``Roofline-TX2'' in the \Fig{fig:roofline-drc}. For the dual compute redundancy, the increase in payload weight lowers the maximum acceleration capability of the UAV, which lowers the roofline (annotated as ``Roofline- 2$\times$ TX2 in \Fig{fig:roofline-drc}), thereby reducing the safe velocity by 33\% compared to the baseline. Thus, there is a trade-off between enabling redundancy and robot operational performance. 

To overcome the drop in operational efficiency of this UAV due to dual compute redundancy, architects can replace the over-provisioned TX2 with an onboard computer with $\frac{1}{5}$\textsuperscript{th} of throughput for DroNet. This will lower the TDP, which will help accommodate two onboard computers within the same power envelope and reduce the payload weight.


\textbf{Takeaway.} 
Dual modular redundancies improve UAVs' safety with the downside of increasing costs and lower performance. To fully understand the consequences of redundancies, there is a need to characterize the effects of additional payload and its impact on a UAV's decision-making rate and physics. F-1 gives us the intuition into various bottlenecks and performance targets to build safe yet efficient redundancy systems.

\subsection{Putting it All Togther: Full System Characterization}
\label{sec:full}


In this case study, we answer the following question:
\textit{Given the choice of several onboard computers, autonomy algorithms, and sensors, how do we systematically characterize and select components to maximize the UAV's safe velocity? How does this selection differ as we change the UAV types?} 


\textbf{UAV Configurations.} We consider several different choices for each UAV.  
For onboard computers, we consider Nvidia TX2, Nvidia AGX, Ras-Pi, and Intel NCS. Similarly, for different UAVs, we consider AscTec Pelican and DJI Spark. Finally for autonomy algorithms, we consider DroNet~\cite{dronet}, TrailNet~\cite{trailnet}, and CAD2RL~\cite{cad2rl}. Using Skyline, we characterize each combination and we share insights from the F-1 model on how we can optimize each of these combinations.



\begin{figure}[t]
\centering
\begin{subfigure}{\linewidth}
        \centering
        \includegraphics[width=\columnwidth,keepaspectratio]{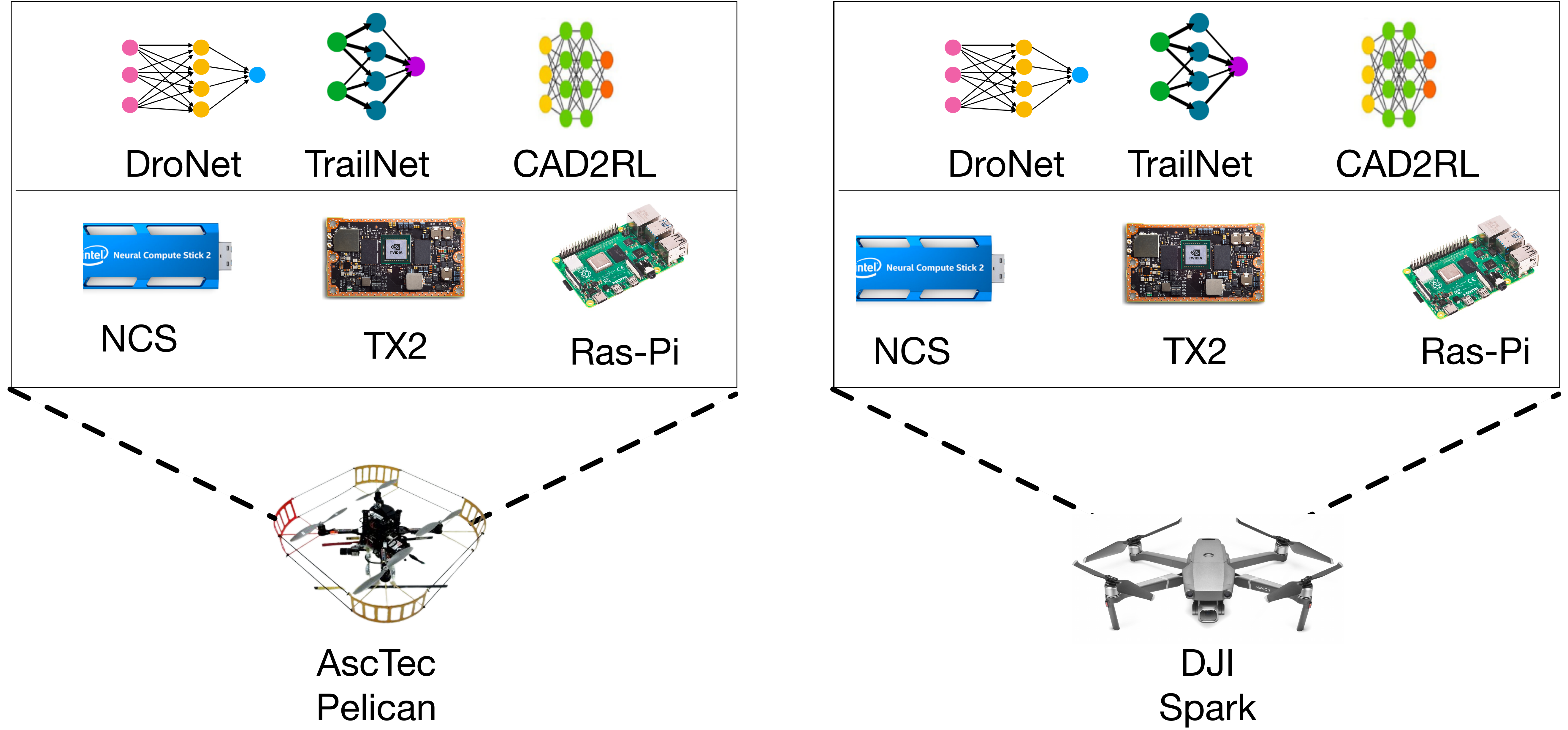}
        \caption{Full UAV system characterization.}
        \label{fig:diff-drones}
        \end{subfigure}
        \begin{subfigure}{\linewidth}
        \centering
        \includegraphics[width=\columnwidth, height=1.5in, keepaspectratio]{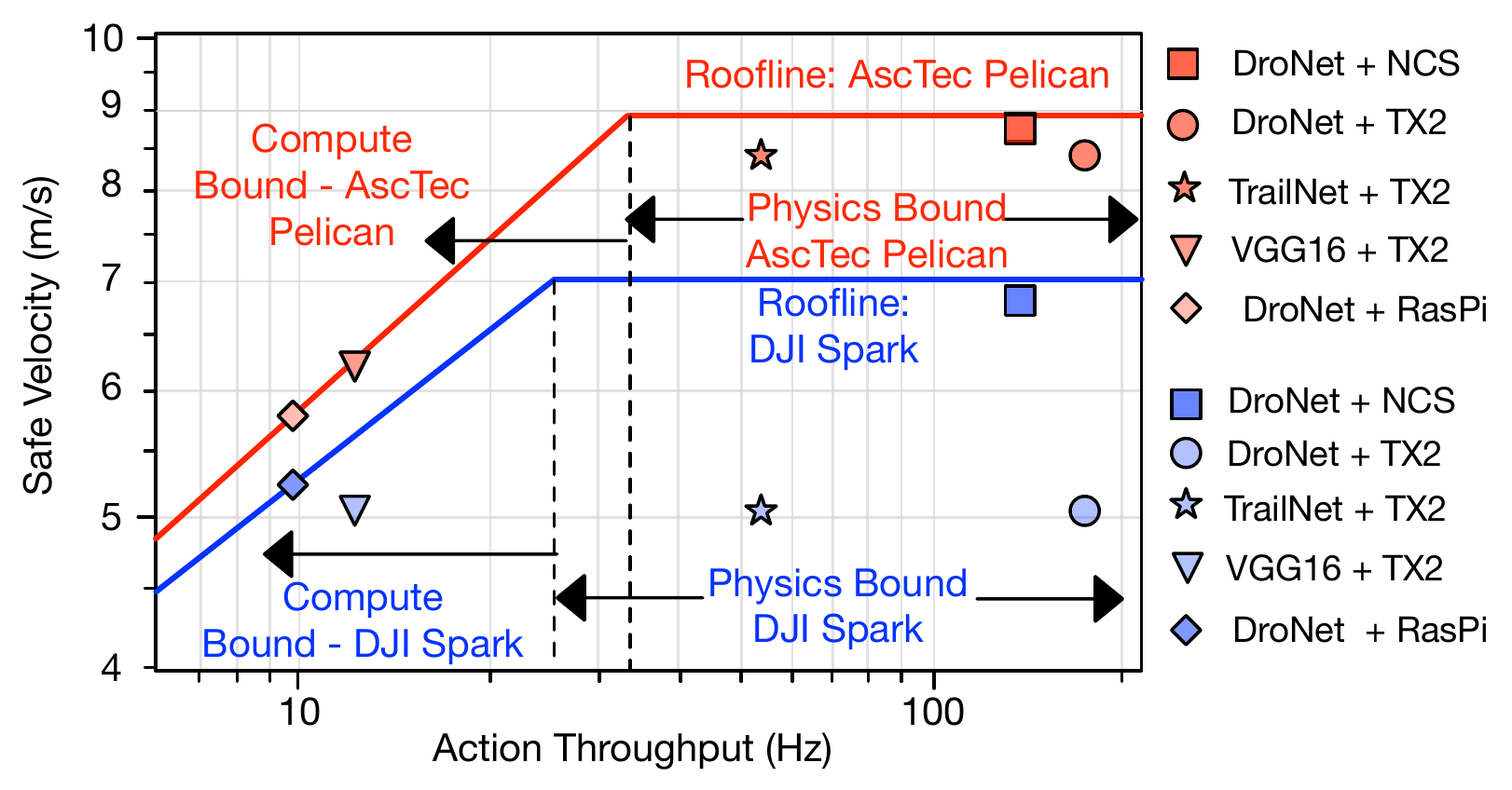}
        \caption{F-1 plot.}
        \label{fig:roofline-full-system}
        \end{subfigure}
    \label{fig:fund-relation}
  \caption{F-1 model to characterize the full UAV system.}
\end{figure}

\textbf{Analysis.} The annotated version of the results are combined and shown in \Fig{fig:roofline-full-system}. Based on the F-1 plots, we can classify these designs as compute-bound or physics-bound. 
Below, we discuss what computer architects can do if the design point falls into the compute-bound/physics-bound category.

\textit{Compute-Bound Designs.} In the compute-bound scenarios, the safe velocity is bounded by the compute throughput of the autonomy algorithm. For designs in this region, computer architects can apply the traditional optimization techniques ranging from micro-architectural to algorithmic optimizations to improve the compute throughput. For instance, in the AscTec Pelican UAV, Ras-Pi4 does not have sufficient computing capability to run autonomy algorithms such as DroNet, TrailNet, and CAD2RL. Therefore, any new architectural optimization for Ras-Pi4 (e.g., building a custom accelerator within the Ras-Pi4 system) would need to improve the compute throughput by 3.3$\times$ for DroNet, 110$\times$ for TrailNet, and 660$\times$ for CAD2RL.

\textit{Physics-Bound Designs.} In physics-bound design, the safe velocity is bounded by the physical dynamics of the UAV rather than the autonomy algorithm or the onboard compute system's performance. If the designs in this region lower the roofline, then computer architects can help alleviate this problem by optimizing for lower TDP. For instance, in the case of DJI Spark with TX2, as an onboard compute running DroNet, it achieves a throughput of 178 Hz. However, the knee-point for this UAV is only 30 Hz suggesting that it is over-provisioned by a factor of 6$\times$. We can trade off this excess performance for a lower TDP (e.g., at a lower clock frequency) in order to reduce its heatsink weight and other board-level components, thus lowering the overall payload weight effect on the UAV.

\textbf{Takeaway.} Ad-hoc selection of UAV components or designing them in isolation impacts UAV's performance. Hence, a systematic methodology is needed to characterize these UAV systems holistically. Skyline tool can give fundamental insights into how various component selections impact the UAV's performance. Moreover, they also provide optimization goals for designing onboard compute for workload targeting UAVs.
\section{Pitfalls in Designing Hardware Accelerators}
\label{sec:discussion}

\begin{figure*}[t]
      \hspace{10pt}
\begin{subfigure}{0.35\columnwidth}
        \includegraphics[width=1\columnwidth,keepaspectratio]{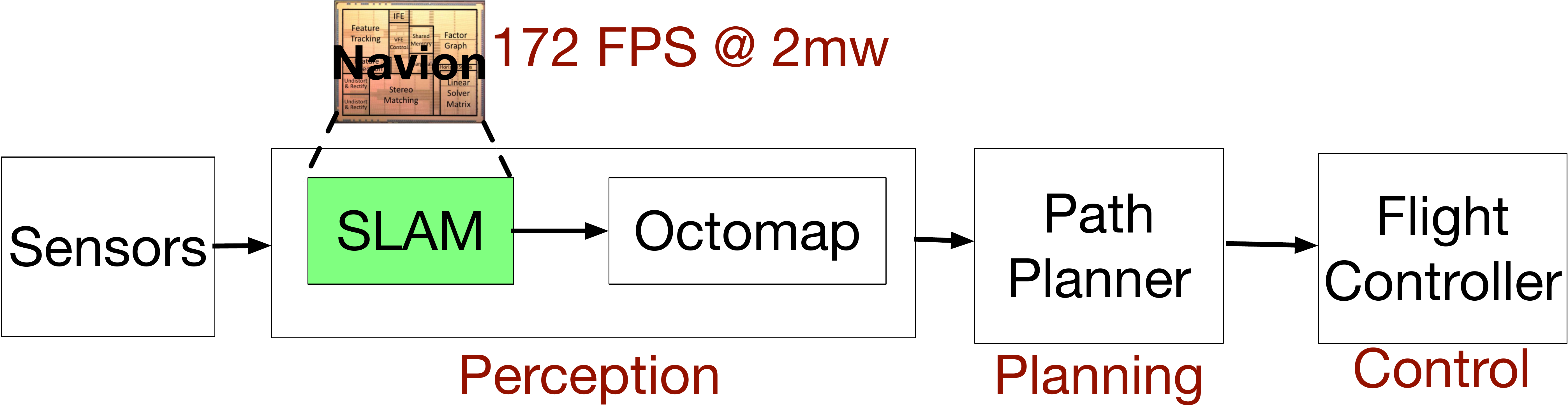}
        \caption{Navion~\cite{navion} evaluation.  }
        \label{fig:navion-flow}
        \end{subfigure}
        \hfill
        \begin{subfigure}{0.25\columnwidth}
        \vspace{-2pt}
        \includegraphics[width=0.85\columnwidth, keepaspectratio]{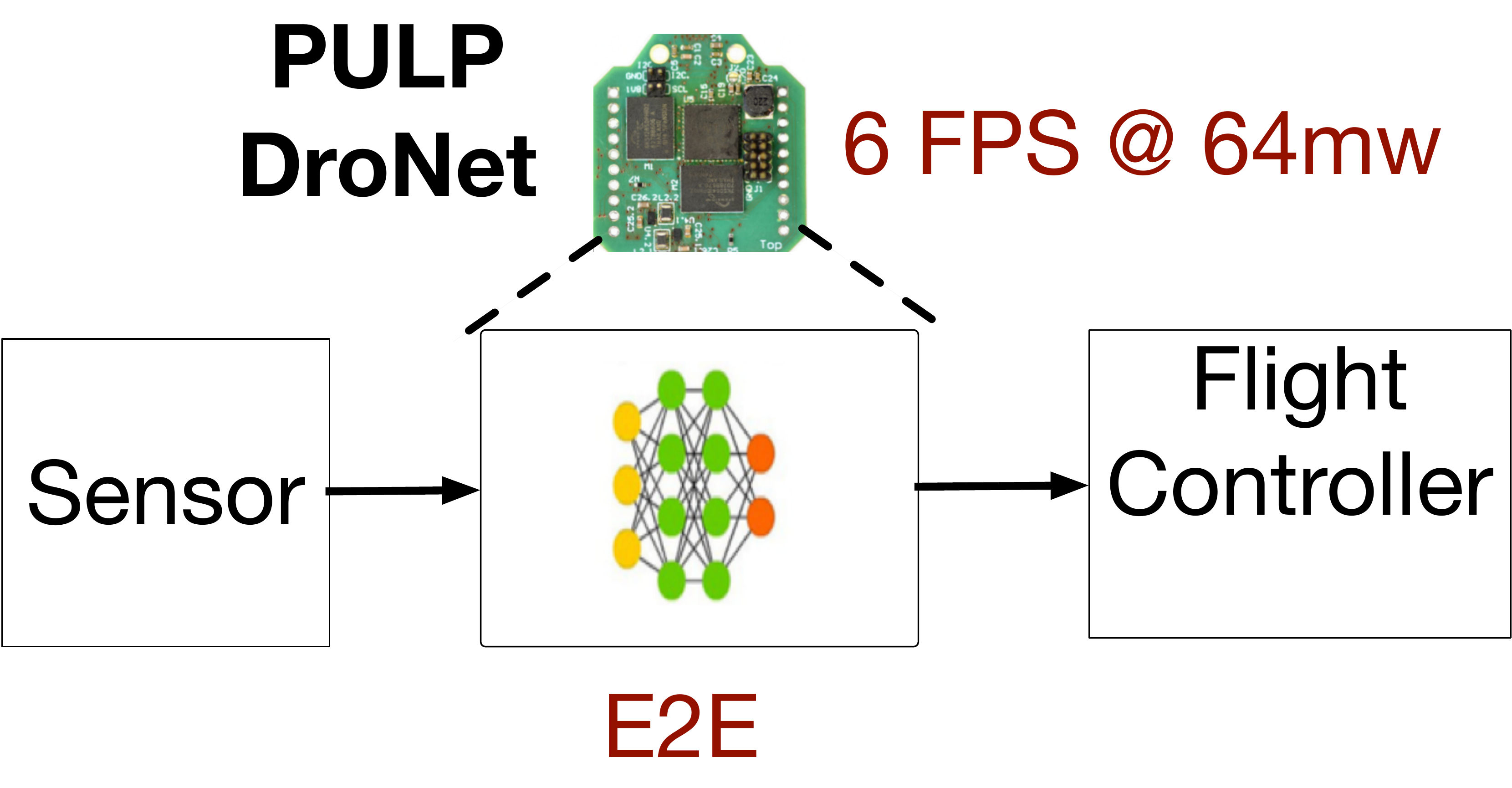}
        \hfill
        \vspace{-1pt}
        \caption{PULP~\cite{pulp-dronet} evaluation. }
        \label{fig:pulp-flow}
        \end{subfigure}
         \begin{subfigure}{0.33\columnwidth}
        \includegraphics[width=0.95\columnwidth, keepaspectratio]{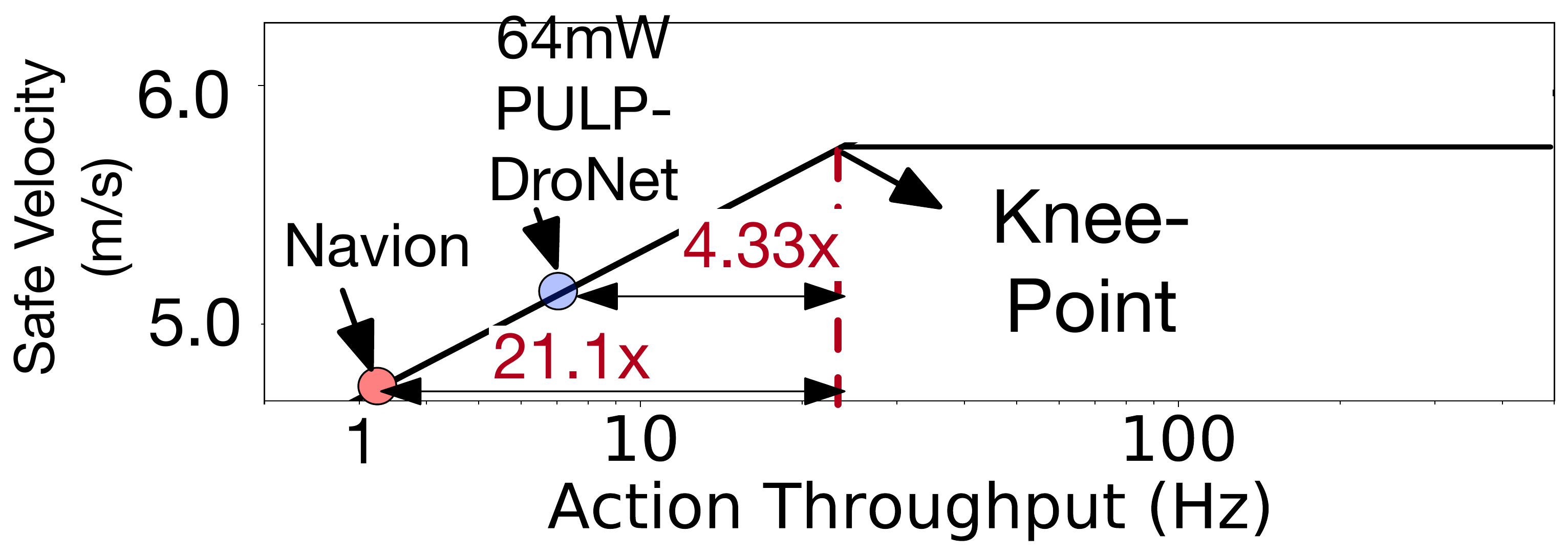}
        \vspace{-1pt}
        \caption{F-1 model. }
        \label{fig:discussion-f1}
        \end{subfigure}
\caption{Evaluation of custom hardware accelerators built specifically for autonomous UAV applications.}
  \label{fig:sensor-study}
\end{figure*}

Thus far, we have presented various case studies to demonstrate the general usefulness of the F-1 model. In this section, we pivot to using the F-1 model to discuss the common pitfalls in designing domain-specific hardware accelerators for UAVs based purely on isolated compute metrics such as `compute throughput' and `low-power.' We consider two popular accelerators, namely Navion~\cite{navion} and PULP-DroNet~\cite{pulp-dronet} that were specifically built for UAVs. We characterize them for a nano-UAV~\cite{nano-uav}. Our characterization demonstrates the need for a systematic approach to understanding the performance requirement needed for a particular UAV rather than designing a high-performance/low-power computer in isolation.


\textbf{Navion.} Navion is a hardware accelerator for visual-inertial odometry in the SPA paradigm, which achieves a throughput of 172 FPS in 2mW. Using Navion in a UAV to achieve full autonomy still requires other algorithms like integration of maps, motion planning, and control. \Fig{fig:navion-flow} shows how the Navion chip will be used in an end-to-end UAV application. For the other stages in the SPA, we characterize the package-delivery application in MAVBench~\cite{mavbench} and replace the SLAM with performance numbers reported in Navion~\cite{navion}. This estimate is still optimistic for Navion, considering it does not perform the loop closure~\cite{hess2016real} in SLAM, whereas in our characterization in MAVBench, the SLAM kernel accounts for loop closure. Briefly, loop closure is the recognition of when a UAV has returned to a previous region that it has mapped. It helps reduce the uncertainty in the estimation of the map.

\textbf{PULP-DroNet.} PULP~\cite{pulp-dronet} is designed for accelerating DroNet~\cite{dronet} for a nano-UAV. It achieves 6 Hz at 64 mW power~\cite{pulp-dronet}. Unlike Navion, PULP can enable end-to-end autonomous operations (full autonomy) in UAVs since DroNet directly operates on raw sensor information to control the UAV. 

\textbf{Analysis.} We characterize the Navion and PULP as onboard compute along with other components of the nano-UAV~\cite{nano-uav}. We also assume the sensor framerate is 60~FPS to ensure we don't fall into sensor-bound scenarios. The F-1 plot for these two configurations are shown in \Fig{fig:discussion-f1}. PULP achieves a throughput of 6~FPS @ 64 mW, which results in a compute-bound scenario since it is to the left of the knee-point in the F-1 roofline plot (see \Fig{fig:discussion-f1}). Therefore, the performance of the PULP hardware accelerator has to be increased by 4.33$\times$ to achieve a peak velocity achievable by this nano-UAV. 

In the case of the Navion chip, even though the SLAM stage achieves an impressive performance of 172 FPS @ 2 mW, integrating into the complete SPA pipeline (\Fig{fig:navion-flow}) increases the overall latency to 810 ms or action throughput of 1.23~Hz. This accelerator also results in a compute-bound scenario since the UAV knee-point throughput is at 26 Hz. Therefore, the end-to-end throughput of the SPA pipeline must be improved by 21.1$\times$ to achieve the peak safe-velocity. Improvements should target building accelerators for mapping~\cite{octomap-acc} and path-planning stages in the SPA pipeline within a similar power envelope. 

\textbf{Takeaway.} Optimizing or designing a specialized compute engine for UAVs based on isolated compute metrics like throughput, low power, or energy efficiency can be misleading, and one needs to consider the compute performance along with the entire autonomy pipeline and other UAV components.  Instead, we recommend using the F-1 model to guide architects' intuitions and perform bottleneck analysis to guide optimization efforts more systematically that translate to UAVs capabilities.

\section{Related Work}

In addition to the related work already discussed in the previous sections, other work specifically in the context of models can be broadly bucketed into compute system models and models that are specifically related to autonomy.

\textbf{Roofline Models.} Roofline models~\cite{roofline} and Gables~\cite{hill2019gables} (which is a roofline model for mobile SoCs with multiple IPs) provide insights into the optimization effort required to maximize the compute throughput for a given workload. These roofline models look at isolated compute throughput and do not include UAVs physics. In contrast, our earlier work~\cite{f-1_cal} introduces the existence of a roofline-like model for UAVs. In this work, we extend the scope of the F-1 model with many case studies and build an interactive web-based tool that shows insights into how much optimization is required (similar to the goals of the roofline model) to maximize the drone's performance by considering compute, sensor, and UAV physics. As we build autonomous vehicles of all sorts and forms, our work shows the importance of considering systems holistically, rather than focusing on isolated components such as hardware acceleration in a vacuum as that can lead to suboptimal designs.

\textbf{Models for Nano-Satellites.} Cote~\cite{orbital-computing} considers orbital mechanics for computer design in nano-satellites. They operate at a different regime than quadcopters and thus are not readily comparable or usable for drones. Nonetheless, they are relevant to our work since they have the same fundamental aspects of considering the systems holistically.



\section{Conclusion}

The F-1 model is an early-stage performance model to systematically characterize and identify bottlenecks when designing onboard compute for autonomous UAVs while considering the full view of the system (i.e., sensor, compute, and physics). Using the model, we can understand how much a custom domain-specific hardware widget improves a given UAV's safe velocity. We believe that the model can be used for automated design space exploration and aid with generating an optimal domain-specific  architecture best suited for a UAV.

\section{Acknowledgements}
The authors thank Magnus Sj{\"a}lander and the other anonymous reviewers for their valuable feedback. The work was sponsored in part by IARPA award 2022-21100600004.

\bibliographystyle{ieeetr}
\bibliography{ref}

\begin{thebibliography}{10}

\bibitem{Timothy2017}
A.~Timothy, M.~N. Paul, A.~T. Aaron, B.~Joan, and S.~Jeff, ``Drone
  transportation of blood products,'' {\em TRANSFUSION Journal}, vol.~57,
  no.~3, pp.~582--588, 2017.

\bibitem{medical-org}
A.~Momont, ``Ambulance drone.'' \url{https:
  //www.tudelft.nl/en/ide/research/research-labs/applied-
  labs/ambulance-drone/}.

\bibitem{search-and-rescue-org}
J.~Rogers, ``How drones are helping the nepal earthquake relief effort.''
  \url{http://www.foxnews.com/tech/2015/04/30/ how- drones- are- helping-
  nepal- earthquake- relief- effort.html}.

\bibitem{8373043}
M.~{Bacco}, A.~{Berton}, E.~{Ferro}, C.~{Gennaro}, A.~{Gotta}, S.~{Matteoli},
  F.~{Paonessa}, M.~{Ruggeri}, G.~{Virone}, and A.~{Zanella}, ``Smart farming:
  Opportunities, challenges and technology enablers,'' in {\em 2018 IoT
  Vertical and Topical Summit on Agriculture - Tuscany (IOT Tuscany)}, 2018.

\bibitem{wan2021survey}
Z.~Wan, B.~Yu, T.~Y. Li, J.~Tang, Y.~Zhu, Y.~Wang, A.~Raychowdhury, and S.~Liu,
  ``A survey of fpga-based robotic computing,'' {\em IEEE Circuits and Systems
  Magazine}, vol.~21, no.~2, pp.~48--74, 2021.

\bibitem{liu2021robotic}
S.~Liu, Z.~Wan, B.~Yu, and Y.~Wang, ``Robotic computing on fpgas,'' {\em
  Synthesis Lectures on Computer Architecture}, vol.~16, no.~1, 2021.

\bibitem{quad-marketshare-1}
``{Commercial Drone Industry Trends}.''
  \url{https://medium.com/aerial-acuity/commercial-drone-industry-trends-aae2010ff349},
  2020.

\bibitem{quad-marketshare-2}
``Commercial drone market size, share \& trends analysis report by product
  (fixed-wing, rotary blade, hybrid), by application, by end-use, by region,
  and segment forecasts, 2021 - 2028.''
  \url{https://www.grandviewresearch.com/industry-analysis/global-commercial-drones-market},
  2020.

\bibitem{search-and-rescue}
A.~Qiantori, A.~B. Sutiono, H.~Hariyanto, H.~Suwa, and T.~Ohta, ``An emergency
  medical communications system by low altitude platform at the early stages of
  a natural disaster in indonesia,'' {\em J. Med. Syst.}, vol.~36, 2012.

\bibitem{package-org-1}
E.~Weise, ``Amazon delivered its first customer package by drone.''
  \url{https://www.usatoday.com/story/tech/news/2016/12/14/
  amazon-delivered-its-first-customer-package-drone/95401366/}.

\bibitem{surveillance-org}
D.~R.~C. McCullough, ``Unmanned aircraft systems(uas) guidebook in
  development.''
  \url{https://cops.usdoj.gov/html/dispatch/08-2014/uas_guidebook_in_development.asp}.

\bibitem{photographyh-org}
R.~Feltman, ``The future of sports photography: Drones.''
  \url{https://www.theatlantic.com/technology/archive/2014/02/the-future-of-sports-photography-drones/283896/}.

\bibitem{agx-power}
``Deploy ai-powered autonomous machines at scale.''
  \url{https://www.nvidia.com/en-us/autonomous-machines/embedded-systems/jetson-agx-xavier/}.

\bibitem{TX2}
``Nvidia jetson tx2 module.''
  \url{https://developer.nvidia.com/embedded/jetson-tx2}.

\bibitem{movidius-drone}
``Intel movidius vision processing units (vpus).''
  \url{https://www.intel.com/content/www/us/en/products/details/processors/movidius-vpu.html},
  2017.

\bibitem{wan2021energy}
Z.~Wan, Y.~Zhang, A.~Raychowdhury, B.~Yu, Y.~Zhang, and S.~Liu, ``An
  energy-efficient quad-camera visual system for autonomous machines on fpga
  platform,'' in {\em 2021 IEEE 3rd International Conference on Artificial
  Intelligence Circuits and Systems (AICAS)}, pp.~1--4, IEEE, 2021.

\bibitem{darpa}
P.~Root, ``{DARPA FLA}.''
  \url{https://www.darpa.mil/program/fast-lightweight-autonomy}.

\bibitem{vijaykumar-1}
K.~Mohta, K.~Sun, S.~Liu, M.~Watterson, B.~Pfrommer, J.~Svacha, Y.~Mulgaonkar,
  C.~J. Taylor, and V.~Kumar, ``Experiments in fast, autonomous, gps-denied
  quadrotor flight,'' in {\em 2018 IEEE International Conference on Robotics
  and Automation (ICRA)}, IEEE, 2018.

\bibitem{vijaykumar-2}
G.~Loianno, D.~Scaramuzza, and V.~Kumar, ``Special issue on high-speed
  vision-based autonomous navigation of uavs,'' {\em Journal of Field
  Robotics}, vol.~1, no.~1, pp.~1--3, 2018.

\bibitem{droneracing}
S.~Li, M.~M. Ozo, C.~De~Wagter, and G.~C. de~Croon, ``Autonomous drone race: A
  computationally efficient vision-based navigation and control strategy,''
  {\em arXiv preprint arXiv:1809.05958}, 2018.

\bibitem{mavbench}
B.~Boroujerdian, H.~Genc, S.~Krishnan, W.~Cui, M.~Almeida, K.~Mansoorshahi,
  A.~Faust, and V.~J. Reddi, ``Mavbench: Micro aerial vehicle benchmarking,''
  in {\em 51st Annual IEEE/ACM International Symposium on Microarchitecture
  (MICRO)}, pp.~894--907, 2018.

\bibitem{high-speed-drone}
S.~Liu, M.~Watterson, S.~Tang, and V.~Kumar, ``High speed navigation for
  quadrotors with limited onboard sensing,'' in {\em 2016 IEEE International
  Conference on Robotics and Automation (ICRA)}, IEEE, 2016.

\bibitem{dji-spark}
DJI, ``{DJI Spark}.'' \url{https://www.dji.com/spark}, 2020.

\bibitem{roofline}
S.~Williams, A.~Waterman, and D.~Patterson, ``Roofline: an insightful visual
  performance model for multicore architectures,'' {\em CACM}, vol.~52, no.~4,
  2009.

\bibitem{redundancy-2}
``Navio2 overview.''
  \url{https://ardupilot.org/copter/docs/common-navio2-overview.html}, 2020.

\bibitem{pulp-dronet}
D.~Palossi, A.~Loquercio, F.~Conti, E.~Flamand, D.~Scaramuzza, and L.~Benini,
  ``A 64mw dnn-based visual navigation engine for autonomous nano-drones,''
  {\em IEEE Internet of Things Journal}, 2019.

\bibitem{nano-uav}
X.~Zhang, B.~Xian, B.~Zhao, and Y.~Zhang, ``Autonomous flight control of a nano
  quadrotor helicopter in a gps-denied environment using on-board vision,''
  {\em IEEE Transactions on Industrial Electronics}, vol.~62, no.~10,
  pp.~6392--6403, 2015.

\bibitem{simple-models}
M.~D. Hill, ``Three other models of computer system performance,'' {\em CoRR},
  vol.~abs/1901.02926, 2019.

\bibitem{hill2019gables}
M.~Hill and V.~J. Reddi, ``Gables: A roofline model for mobile socs,'' in {\em
  2019 IEEE International Symposium on High Performance Computer Architecture
  (HPCA)}, pp.~317--330, IEEE, 2019.

\bibitem{crazyflie}
``Crazyflie 2.1 product page.''
  \url{https://www.bitcraze.io/products/crazyflie-2-1/ }, 2020.

\bibitem{ucontroller-fc-1}
``Pixhawk 1 flight controller..''
  \url{https://docs.px4.io/v1.9.0/en/flight_controller/pixhawk.html}, 2020.

\bibitem{ucontroller-fc-2}
Get{FPV}, ``All about multirotor drone fpv flight controllers.''
  \url{https://www.getfpv.com/learn/new-to-fpv/all-about-multirotor-fpv-drone-flight-controller/},
  2020.

\bibitem{imu}
M.~Achtelik, T.~Zhang, K.~Kuhnlenz, and M.~Buss, ``Visual tracking and control
  of a quadcopter using a stereo camera system and inertial sensors,'' in {\em
  2009 International Conference on Mechatronics and Automation},
  pp.~2863--2869, IEEE, 2009.

\bibitem{1khz-control}
D.~Gurdan, J.~Stumpf, M.~Achtelik, K.-M. Doth, G.~Hirzinger, and D.~Rus,
  ``Energy-efficient autonomous four-rotor flying robot controlled at 1 khz,''
  in {\em Proceedings 2007 IEEE International Conference on Robotics and
  Automation}, pp.~361--366, IEEE, 2007.

\bibitem{koch2019neuroflight}
W.~Koch, R.~Mancuso, and A.~Bestavros, ``Neuroflight: Next generation flight
  control firmware,'' {\em arXiv preprint arXiv:1901.06553}, 2019.

\bibitem{rusu20113d}
R.~B. Rusu and S.~Cousins, ``3d is here: Point cloud library (pcl),'' in {\em
  2011 IEEE international conference on robotics and automation}, 2011.

\bibitem{elfes1989using}
A.~Elfes, ``Using occupancy grids for mobile robot perception and navigation,''
  {\em Computer}, vol.~22, no.~6, pp.~46--57, 1989.

\bibitem{dissanayake2001solution}
M.~G. Dissanayake, P.~Newman, S.~Clark, H.~F. Durrant-Whyte, and M.~Csorba, ``A
  solution to the simultaneous localization and map building (slam) problem,''
  {\em IEEE Transactions on robotics and automation}, vol.~17, no.~3, 2001.

\bibitem{gao2021ielas}
T.~Gao, Z.~Wan, Y.~Zhang, B.~Yu, Y.~Zhang, S.~Liu, and A.~Raychowdhury,
  ``ielas: An elas-based energy-efficient accelerator for real-time stereo
  matching on fpga platform,'' in {\em 2021 IEEE 3rd International Conference
  on Artificial Intelligence Circuits and Systems (AICAS)}, IEEE, 2021.

\bibitem{rrt}
S.~Karaman and E.~Frazzoli, ``Sampling-based algorithms for optimal motion
  planning,'' {\em The international journal of robotics research}, vol.~30,
  no.~7, pp.~846--894, 2011.

\bibitem{motion-planning-survey}
D.~{Gonzalez}, J.~{Perez}, V.~{Milanes}, and F.~{Nashashibi}, ``A review of
  motion planning techniques for automated vehicles,'' {\em IEEE Transactions
  on Intelligent Transportation Systems}, pp.~1135--1145, 2016.

\bibitem{e2e-nvidia}
M.~Bojarski, D.~Del~Testa, D.~Dworakowski, B.~Firner, B.~Flepp, P.~Goyal, L.~D.
  Jackel, M.~Monfort, U.~Muller, J.~Zhang, {\em et~al.}, ``End to end learning
  for self-driving cars,'' {\em arXiv preprint arXiv:1604.07316}, 2016.

\bibitem{dronet}
A.~Loquercio, A.~I. Maqueda, C.~R. Del-Blanco, and D.~Scaramuzza, ``Dronet:
  Learning to fly by driving,'' {\em IEEE Robotics and Automation Letters},
  vol.~3, no.~2, pp.~1088--1095, 2018.

\bibitem{trail-net}
N.~Smolyanskiy, A.~Kamenev, J.~Smith, and S.~Birchfield, ``Toward low-flying
  autonomous mav trail navigation using deep neural networks for environmental
  awareness,'' in {\em 2017 IEEE/RSJ International Conference on Intelligent
  Robots and Systems (IROS)}, pp.~4241--4247, IEEE, 2017.

\bibitem{cad2rl}
F.~Sadeghi and S.~Levine, ``Cad2rl: Real single-image flight without a single
  real image,'' {\em arXiv preprint arXiv:1611.04201}, 2016.

\bibitem{qt-opt}
D.~Kalashnikov, A.~Irpan, P.~Pastor, J.~Ibarz, A.~Herzog, E.~Jang, D.~Quillen,
  E.~Holly, M.~Kalakrishnan, V.~Vanhoucke, {\em et~al.}, ``Qt-opt: Scalable
  deep reinforcement learning for vision-based robotic manipulation,'' {\em
  arXiv preprint arXiv:1806.10293}, 2018.

\bibitem{quarl}
S.~Krishnan, S.~Chitlangia, M.~Lam, Z.~Wan, A.~Faust, and V.~J. Reddi,
  ``Quantized reinforcement learning {(QUARL)},'' {\em CoRR},
  vol.~abs/1910.01055, 2019.

\bibitem{airlearning}
S.~Krishnan, B.~Boroujerdian, W.~Fu, A.~Faust, and V.~J. Reddi, ``Air learning:
  a deep reinforcement learning gym for autonomous aerial robot visual
  navigation,'' {\em Machine Learning}, vol.~110, pp.~2501--2540, 2021.

\bibitem{anwar2020autonomous}
A.~Anwar and A.~Raychowdhury, ``Autonomous navigation via deep reinforcement
  learning for resource constraint edge nodes using transfer learning,'' {\em
  IEEE Access}, vol.~8, pp.~26549--26560, 2020.

\bibitem{fov}
``Field of view.'' \url{https://en.wikipedia.org/wiki/Field_of_view}, 2020.

\bibitem{prm-rl}
A.~Faust, O.~Ramirez, M.~Fiser, K.~Oslund, A.~G. Francis, J.~Davidson, and
  L.~Tapia, ``{PRM-RL:} long-range robotic navigation tasks by combining
  reinforcement learning and sampling-based planning,'' {\em CoRR},
  vol.~abs/1710.03937, 2017.

\bibitem{trailnet}
N.~Smolyanskiy, A.~Kamenev, J.~Smith, and S.~Birchfield, ``Toward low-flying
  autonomous {MAV} trail navigation using deep neural networks for
  environmental awareness,'' in {\em IEEE/RSJ International Conference on
  Intelligent Robots and Systems (IROS)}, pp.~4241--4247, 2017.

\bibitem{krishnan2021autosoc}
S.~Krishnan, T.~Tambe, Z.~Wan, and V.~J. Reddi, ``Autosoc: Automating
  algorithm-soc co-design for aerial robots,'' {\em arXiv preprint
  arXiv:2109.05683}, 2021.

\bibitem{heat-sink}
``Heat sink size calculator.''
  \url{https://celsiainc.com/resources/calculators/heat-sink-size-calculator/}.
\newblock (Accessed on 01/29/2020).

\bibitem{wan2021analyzing}
Z.~Wan, A.~Anwar, Y.-S. Hsiao, T.~Jia, V.~J. Reddi, and A.~Raychowdhury,
  ``Analyzing and improving fault tolerance of learning-based navigation
  systems,'' in {\em 2021 58th ACM/IEEE Design Automation Conference (DAC)},
  pp.~841--846, IEEE, 2021.

\bibitem{wan2022frlfi}
Z.~Wan, A.~Anwar, A.~Mahmoud, T.~Jia, Y.-S. Hsiao, V.~J. Reddi, and
  A.~Raychowdhury, ``Frl-fi: Transient fault analysis for federated
  reinforcement learning-based navigation systems,'' in {\em 2022 Design,
  Automation \& Test in Europe Conference \& Exhibition (DATE)}, 2022.

\bibitem{bannon2019computer}
P.~Bannon, G.~Venkataramanan, D.~D. Sarma, and E.~Talpes, ``Computer and
  redundancy solution for the full self-driving computer,'' in {\em 2019 IEEE
  Hot Chips 31 Symposium (HCS)}, IEEE Computer Society, 2019.

\bibitem{mehmed2020monitor}
A.~Mehmed, M.~Antlanger, and W.~Steiner, ``The monitor as key architecture
  element for safe self-driving cars,'' in {\em 2020 50th Annual IEEE-IFIP
  International Conference on Dependable Systems and Networks-Supplemental
  Volume (DSN-S)}, pp.~9--12, IEEE, 2020.

\bibitem{navion}
A.~Suleiman, Z.~Zhang, L.~Carlone, S.~Karaman, and V.~Sze, ``Navion: A 2-mw
  fully integrated real-time visual-inertial odometry accelerator for
  autonomous navigation of nano drones,'' {\em IEEE Journal of Solid-State
  Circuits}, vol.~54, no.~4, pp.~1106--1119, 2019.

\bibitem{hess2016real}
W.~Hess, D.~Kohler, H.~Rapp, and D.~Andor, ``Real-time loop closure in 2d lidar
  slam,'' in {\em 2016 IEEE international conference on robotics and automation
  (ICRA)}, pp.~1271--1278, IEEE, 2016.

\bibitem{octomap-acc}
T.~Jia, E.-Y. Yang, Y.-S. Hsiao, J.~Cruz, D.~Brooks, G.-Y. Wei, and V.~J.
  Reddi, ``Omu: A probabilistic 3d occupancy mapping accelerator for real-time
  octomap at the edge,'' in {\em Design, Automation and Test in Europe (DATE)},
  Mar 2022.

\bibitem{f-1_cal}
S.~Krishnan, Z.~Wan, K.~Bhardwaj, P.~Whatmough, A.~Faust, G.-Y. Wei, D.~Brooks,
  and V.~J. Reddi, ``The sky is not the limit: A visual performance model for
  cyber-physical co-design in autonomous machines,'' {\em IEEE Computer
  Architecture Letters}, 2020.

\bibitem{orbital-computing}
B.~Denby and B.~Lucia, ``Orbital edge computing: Nanosatellite constellations
  as a new class of computer system,'' in {\em Proceedings of the Twenty-Fifth
  International Conference on Architectural Support for Programming Languages
  and Operating Systems}, pp.~939--954, 2020.

\end{thebibliography}

\end{document}